\documentclass[10pt,twocolumn,letterpaper]{article}

\usepackage{iccv}
\usepackage{times}
\usepackage{epsfig}
\usepackage{graphicx}
\usepackage{amsmath}
\usepackage{amssymb}

\usepackage{verbatim}
\usepackage{booktabs}       
\usepackage{amsfonts}       
\usepackage{nicefrac}       
\usepackage{microtype}      
\usepackage{bm}             
\usepackage{mathtools}      
\usepackage{comment}
\usepackage{subfigure} 
\usepackage{pdflscape}
\usepackage{multirow}
\usepackage{soul}
\usepackage{authblk}

\usepackage[pagebackref=true,breaklinks=true,letterpaper=true,colorlinks,bookmarks=false]{hyperref}

\iccvfinalcopy 

\def\iccvPaperID{4549} 

\ificcvfinal\fi
\begin{document}

\title{Bayes-Factor-VAE: Hierarchical Bayesian Deep Auto-Encoder Models for\\
Factor Disentanglement\vspace{-1.5em}}
\author[1]{Minyoung Kim
}
\author[1,2]{Yuting Wang
}
\author[2]{Pritish Sahu
}
\author[1,2]{Vladimir Pavlovic
}
\affil[1]{Samsung AI Center, Cambridge, UK}
\affil[2]{Dept. of Computer Science, Rutgers University, NJ, USA}
\affil[ ]{\normalsize{\texttt{mikim21@gmail.com}, \texttt{\{yw632,ps851,vladimir\}@cs.rutgers.edu}, \texttt{v.pavlovic@samsung.com}}
\vspace{-1.5em}
}


\maketitle

\begin{abstract}
We propose a family of novel hierarchical Bayesian deep auto-encoder models capable of identifying disentangled factors of variability in data. While many recent attempts at factor disentanglement have focused on sophisticated learning objectives within the VAE framework, their choice of a standard normal as the latent factor prior is both suboptimal and detrimental to performance.  Our key observation is that the disentangled latent variables responsible for major sources of variability, the {\em relevant factors}, can be more appropriately modeled using long-tail distributions. The typical Gaussian priors are, on the other hand, better suited for modeling of nuisance factors.  
Motivated by this, we extend the VAE to a hierarchical Bayesian model by introducing hyper-priors on the variances of Gaussian latent priors,  mimicking an infinite mixture, while maintaining tractable learning and inference of the traditional VAEs.  This analysis signifies the importance of partitioning and treating in a different manner the latent dimensions corresponding to relevant factors and nuisances. 
Our proposed models, dubbed Bayes-Factor-VAEs, are shown to outperform existing methods both quantitatively and qualitatively in terms of latent disentanglement across several challenging benchmark tasks. 
\end{abstract}

\section{Introduction}\label{sec:intro}

Data, such as images or videos, are inherently high-dimensional, a result of interactions of many complex factors such as lighting, illumination, geometry, etc.  Identifying those factors and their intricate interplay is the key not only to explaining the source of variability in the data but also to efficiently representing the same data for subsequent analysis, classification, or even re-synthesis.  To tackle this problem, deep factor models such as the VAE~\cite{vae14} have been proposed to principally, mathematically concisely, and computationally efficiently model the nonlinear generative relationship between the ambient data and the latent factors.  

However, solely identifying some factors beyond the sources of variability is not sufficient; it is ultimately desirable that the identified factors also be {\em disentangled}. Although there are several different, sometimes opposing, views of disentanglement~\cite{bengio_disent,def_disent}, the most commonly accepted definition aligns with the notion of {\em apriori independence}, where each aspect of independent variability in data is exclusively sourced in one latent factor. Identifying these disentangled factors will then naturally lead to an effective, succinct representation of the data.  In this paper we aim to solve this disentangled representation learning task in the most challenging, {\em unsupervised} setting, with no auxiliary information, such as labels, provided during the learning process. 

While there have been considerable recent efforts to solve the latent disentanglement problem~\cite{infogan16,aae16,beta_vae17,nica17,dip_vae18,factor_vae18,tcvae}, 
most prior approaches have failed to produce satisfactory solutions. One fundamental reason for this is their inadequate treatment of the key factors supporting the disentanglement, which have in prior works been almost universally tied to i.i.d.\ Gaussian priors. In contrast, to accomplish high-quality disentanglement {\em one needs to distinguish, and treat separately, the relevant latent variables, responsible for principal variability in the data, from the nuisance sources of minor variation.  Specifically, the relevant factors may exhibit non-Gaussian, long-tail behavior, which discerns them from statistically independent Gaussian nuisances.} We will detail and justify this requirement in \autoref{sec:analysis}.



Our goal in this paper is to develop principled factor disentanglement algorithms that meet this requirement. In particular, we propose three different hierarchical Bayesian models that place hyper-priors on the parameters of the latent prior. This effectively mimics employing infinite mixtures while maintaining tractable learning and inference of traditional VAEs. We begin with a brief background on VAEs, describe our motivation and requirement to achieve the disentanglement in a principled way (\autoref{sec:analysis}), followed by the definition of specific models (\autoref{sec:bfvae}). 

\textbf{Background.} 
We denote by ${\bf x}\in \mathbb{R}^D$ the observation (e.g., image) and by ${\bf z}\in \mathbb{R}^d$ the underlying latent vector. The variational auto-encoder (VAE)~\cite{vae14} is a deep probabilistic model that represents the joint distribution as:
\begin{eqnarray}
\vspace{-0.5em}
p({\bf z}) &=& \mathcal{N}({\bf z}; {\bf 0}, {\bf I}), \label{eq:p_z} \\
p_\theta({\bf x}|{\bf z}) &=& p({\bf x}; \theta({\bf z})), \label{eq:p_x_given_z}
\vspace{-0.5em}
\end{eqnarray}
where $p({\bf x}; \theta({\bf z}))$ is a density model with the parameters $\theta({\bf z})$ whose likelihood can be tractably computed (e.g., Gaussian or Bernoulli), and $\theta({\bf z})$ is the output of a deep model with its own weight parameters. In the unsupervised setting, with ambient data $\{{\bf x}^n\}_{n=1}^N$, the model can be learned by the MLE, i.e., maximizing $\sum_{n=1}^N \log p({\bf x}^n)$. This requires posterior inference $p({\bf z}|{\bf x})$, but as the exact inference is intractable, the VAE adopts the variational technique: approximate $p({\bf z}|{\bf x}) \approx q_\nu({\bf z}|{\bf x})$, where 
$q_\nu({\bf z}|{\bf x}) = q({\bf z}; \nu({\bf x}))$ is a freely chosen tractable density 
with parameters modeled by deep model $\nu({\bf x})$. A typical choice, assumed throughout the paper, is independent Gaussian, 
\begin{equation}
q_\nu({\bf z}|{\bf x}) 
  = \prod_{j=1}^d \mathcal{N}(z_j; m_j({\bf x}), s_j({\bf x})^2),
\label{eq:q_gaussian} 
\end{equation}
where $\nu({\bf x}) = \{ m_j({\bf x}), s_j({\bf x}) \}_{j=1}^d$ for some deep networks $m_j({\bf x})$ and $s_j({\bf x})$.
%
The negative data log-likelihood admits the following as its upper bound,
\begin{equation}
  \textrm{Rec}(\theta,\nu) + \mathbb{E}_{p_d({\bf x})} \big[ 
     \textrm{KL}( q_\nu({\bf z}|{\bf x}) || p({\bf z}) ) \big],
\label{eq:obj_vae}
\end{equation}
which we minimize wrt $\theta(\cdot)$ and $\nu(\cdot)$.  Here, $p_d({\bf x})$ is the empirical data distribution of $\{{\bf x}^n\}_{n=1}^N$, and 
\begin{equation}
\textrm{Rec}(\theta,\nu) = 
  - \mathbb{E}_{p_d({\bf x})} \big[ E_{q_\nu({\bf z}|{\bf x})} [ \log p_\theta({\bf x}|{\bf z}) ] \big]
\label{eq:recon_loss}
\end{equation}
is the negative expected log-likelihood, identical to the reconstruction loss.

\section{Our Motivation}\label{sec:analysis}

Although minimizing \eqref{eq:obj_vae} can yield a model that faithfully explains the observations, the learned model does not necessarily exhibit {\em disentanglement of latent factors}. In this section, we begin with a common notion of latent disentanglement\footnote{While there is no universal definition, the one we use shares the main concepts with other definitions, including the recent symmetric transformation view~\cite{def_disent}.}, and consider a semi-parametric extension of VAE to derive a principled objective function to achieve latent disentanglement under this notion. Our analysis also suggests to discriminate relevant latent variables from nuisances, and separately treat the two. 

\textbf{Notion of Disentanglement.} 
Consider a set of aspects that can be observed in ${\bf x}$, where the value of each aspect varies independently from the others in the data. In the facial image data, for instance, we typically observe images where the variability of each aspect, say (\texttt{pose}, \texttt{gender}, \texttt{facial expression}), is independent from the others (e.g., the distribution of \texttt{pose} variability in images is the same regardless of \texttt{gender} or \texttt{expression}).  We then say the latent vector ${\bf z}$ is disentangled if each variable $z_j$ is statistically correlated with only a single aspect, exclusive from other $z_{-j}$. That is, varying $z_j$ while fixing $z_{-j}$, results in the exclusive variation of the $j$-th aspect in ${\bf x}$. 

\textbf{Relevant vs.~Nuisance Variables.}
It is natural to assume the exact number of meaningful aspects is a priori unknown, but a sufficiently large upper bound $d$ may be known. Only some variables in ${\bf z}$ will have correspondence to aspects, with the rest attributed to nuisance effects (e.g., acting as a conduit to the data generation process). We thus partition the latent dimensions into two disjoint subsets, ${\bf R}$ (relevant) and ${\bf N}$ (nuisance), ${\bf R} \cup {\bf N} = \{1,\dots,d\}$ and ${\bf R} \cap {\bf N} = \emptyset$. 
Formally, index $j$ is said to be relevant ($j\in{\bf R}$) if $z_j$ and ${\bf x}$ are statistically {\em dependent} 
and $j$ is called nuisance ($j\in{\bf N}$) if $z_j$ and ${\bf x}$ are statistically {\em independent}. 
Analysis in this section assumes {\em known} ${\bf R}$ and ${\bf N}$. 


The above notion implies the latent variables $z_j$'s be apriori independent of each other, in agreement with the goals and framework of the {\em independent component analysis} (ICA)~\cite{ica_book}, the task of blind separation of statistically independent sources. In particular, our derivation is based on the {\em semi-parametric} view~\cite{semi_param_book,kernel_ica}, in which the only assumption made is that of a fully factorized $p({\bf z})$, with no restrictions on the choice of the density $p({\bf z})$. 

For ease of exposition, we consider a {\em deterministic} decoder/encoder pair, ${\bf x} = \textrm{dec}_\theta({\bf z})$ and ${\bf z} = \textrm{enc}_\nu({\bf x})$ with parameters $\theta$ and $\nu$, constrained to be the inverses of each other, $\textrm{enc}_\nu(\cdot) = \textrm{dec}_\theta^{-1}(\cdot)$. 
In the semi-parametric ICA, we seek to solve the MLE problem:
\begin{equation}
\min_{p({\bf z}), \theta} \textrm{KL}\left( p_d({\bf x}) || p_\theta({\bf x}) \right) \ \ \ \
    \textrm{s.t.} \ \ p({\bf z}) = \prod_{j=1}^d p(z_j),
\label{eq:ica_ml}
\end{equation}
where $p_\theta({\bf x})$ is the density derived from ${\bf x} = \textrm{dec}_\theta({\bf z})$, with ${\bf z} \sim p({\bf z})$.  The latent prior $p({\bf z})$ is now a part of our model to be learned, instead of being fixed as in VAE. We let $p({\bf z})$ be of free form (semi-parametric) but fully factorized, the key to the latent disentanglement. 

Directly optimizing \eqref{eq:ica_ml} is intractable, and we solve it in the ${\bf z}$ space. Using the fact that KL divergence is invariant to invertible transformations\footnote{See Supplement for the proof.}, we have:
\begin{equation}
\textrm{KL}( p_d({\bf x}) || p_\theta({\bf x}) ) =
  \textrm{KL}( q_\nu({\bf z}) || p({\bf z}) ),
\label{eq:kl_approx_det}
\end{equation}
where $q_\nu({\bf z})$ is the density of ${\bf z} = \textrm{enc}_\nu({\bf x})$ with ${\bf x} \sim p_d({\bf x})$.  Our original problem \eqref{eq:ica_ml} then becomes:
\begin{equation}
\min_{p({\bf z}), \nu} 
  \textrm{KL}_{\bf z} := \textrm{KL}\Bigg( q_\nu({\bf z}) \bigg\| \prod_{j=1}^d p(z_j) \Bigg) 
\label{eq:ica_ml2} 
\end{equation}

In case when the encoder/decoder pair becomes {\em stochastic} \eqref{eq:p_x_given_z} and \eqref{eq:q_gaussian}, three modifications are needed: i) stochastic inverse\footnote{Such that $\theta$ and $\nu$ minimize the reconstruction loss $\textrm{Rec}(\theta,\nu)$ \eqref{eq:recon_loss}.
}, 
ii) the invariance of KL \eqref{eq:kl_approx_det} turns into an approximation, and iii) $q_\nu({\bf z})$ is defined as:
\begin{equation}
q_\nu({\bf z}) = \mathbb{E}_{p_d({\bf x})} \Big[ q_\nu({\bf z}|{\bf x}) \Big]
    = \frac{1}{N} \sum_{n=1}^N q_\nu({\bf z}|{\bf x}^n),
\label{eq:q_z}
\end{equation}
a well known quantity in the recent disentanglement literature, dubbed {\em aggregate posterior}.  

Further imposing the independence constraint for the nuisance variables, our optimization problem becomes:
\begin{equation}
\min_{p({\bf z}),\nu}
  \textrm{KL}_{\bf z}
\ \ \ \textrm{s.t.} \ \ q_\nu(z_j|{\bf x}) = q_\nu(z_j) \ \ \forall {\bf x}, \ \ j \in {\bf N}, \label{eq:ica_ml2_conN} 
\end{equation}
where $q_\nu(z_j|{\bf x})$ and $q_\nu(z_j)$ are marginals from $q_\nu({\bf z}|{\bf x})$ and $q_\nu({\bf z})$, respectively. We will often omit the subscript $\nu$ in notation. It is not difficult to see that the objective $\textrm{KL}_{\bf z}$ in \eqref{eq:ica_ml2} and \eqref{eq:ica_ml2_conN} can be decomposed as follows (see Supplement): 
\begin{equation}
\textrm{KL}_{\bf z} =
\textrm{TC} + 
    \sum_{j\in{\bf R}} \textrm{KL}( q(z_j) || p(z_j) ) + 
    \sum_{j\in{\bf N}} \textrm{KL}( q(z_j) || p(z_j) ) 
\label{eq:kl_identity_general2} 
\end{equation}
where $\textrm{TC}$ is the {\em total correlation}, a measure of the degree of factorization of $q({\bf z})$: 
\begin{equation}
\textrm{TC} := \textrm{KL}\Bigg( q({\bf z}) \bigg\|  \prod_{j=1}^d q(z_j) \Bigg).
\label{eq:tc}
\end{equation}

With the freedom to choose $p({\bf z})$ and $\nu$ (of $q_\nu({\bf z}|{\bf x})$) to minimize $\textrm{KL}_{\bf z}$ within the constraint \eqref{eq:ica_ml2_conN}, we tackle the last two terms in \eqref{eq:kl_identity_general2} individually. 

\noindent\textbf{$\mathbf{3^{rd}}$ Term.} 
For nuisance $z_j$, to satisfy the constraint \eqref{eq:ica_ml2_conN}, we have $q(z_j|{\bf x}) = \mathcal{N}(z_j; m_j, s_j^2)$ for some fixed $m_j$ and $s_j$. 
Then $q(z_j) := \int q(z_j|{\bf x}) p_d({\bf x}) d{\bf x} = \mathcal{N}(z_j; m_j, s_j^2)$, allowing one to choose a Gaussian prior $p(z_j)=\mathcal{N}(z_j; 0, 1)$, leading to $m_j=0$, $s_j=1$, vanishing the KL. 

\noindent\textbf{$\mathbf{2^{nd}}$ Term.} For $z_j$ a relevant factor variable, $z_j$ and ${\bf x}$ should not be independent, thus $q(z_j)$ is a Gaussian mixture with heterogeneous components $q(z_j|{\bf x})$.  The VAE's Gaussian prior $p(z_j)=\mathcal{N}(z_j; 0, 1)$ implies that the divergence can never be made to vanish in general.  To remedy this, one either i) chooses $p(z_j)$ different from $\mathcal{N}(0,1)$ (potentially, non-Gaussian), or ii) retains a Gaussian prior but lets the mean and variance of $p(z_j)$ be flexibly chosen, perhaps differently over $j\in{\bf R}$, to maximally diminish this KL divergence. 
The former approach may raise a nontrivial question of which prior to choose\footnote{One may employ a flexible model for $p(z_j)$, e.g., a finite mixture approximation or the VampPrior~\cite{vampprior}. However, this may lead to overfitting; see our empirical study in \autoref{sec:quantitative}.}. Instead, we propose a solution that builds a hierarchical Bayesian prior of $p(z_j)$ and infers the posterior (\autoref{sec:bfvae1} and~\ref{sec:bfvae2}). 
In this strategy, we regard the variances of Gaussian $p(z_j)$ as parameters to be learned, and minimize $\textrm{KL}( q(z_j) || p(z_j) )$ wrt the VAE parameters as well as the prior variances (\autoref{sec:bfvae0}). 

\noindent\textbf{Learning Objective.} Based on the above analysis, the overall learning goal can be defined as:
\begin{align}
&\min_{\theta,\nu,p({\bf z})} \ 
    \textrm{Rec}(\theta,\nu) + 
    \mathbb{E}_{p_d({\bf x})}[\textrm{KL}( q(z_j|{\bf x}) || p(z_j) )] + 
    \gamma \textrm{TC}& \nonumber \\
&\ \ \ \ \ \textrm{s.t.} \ \ \ \ p(z_j) = \mathcal{N}(z_j; 0, 1) \ \ \textrm{for} \ \ j \in {\bf N}, \label{eq:obj_disent}&
\end{align}
where we include $\textrm{Rec}(\theta,\nu)$ of \eqref{eq:recon_loss} to impose the stochastic inverse, and replace the difficult-to-evaluate $\textrm{KL}( q(z_j) || p(z_j) )$ by the expected KL, an upper bound\footnote{See Supplement for the proof.}  
admitting a closed form. The TC term will be estimated through its density ratio proxy, using an adversarial  discriminator similarly as~\cite{factor_vae18}, where its impact is controlled by $\gamma$. 

Our learning objective in \eqref{eq:obj_disent} is similar to those of recent disentanglement algorithms (see \autoref{sec:related}) in that the VAE loss is augmented with the additional loss of independence of latent variables, such as the TC term. However, a key distinction is our separate treatment of relevant and nuisance variables, with the additional aim to learn a non-Gaussian relevant variable prior $p(z_j)$. The optimization \eqref{eq:obj_disent} assumes a known relevance partition ${\bf R}$ and ${\bf N}$. In the next section we will deal with how to learn this partition automatically from data, either implicitly (\autoref{sec:bfvae0} and~\ref{sec:bfvae1}) or explicitly (\autoref{sec:bfvae2}) via hierarchical Bayesian treatment.

\begin{figure}
\begin{center}
\includegraphics[trim = 1mm 1mm 1mm 1mm, clip, scale=0.255]{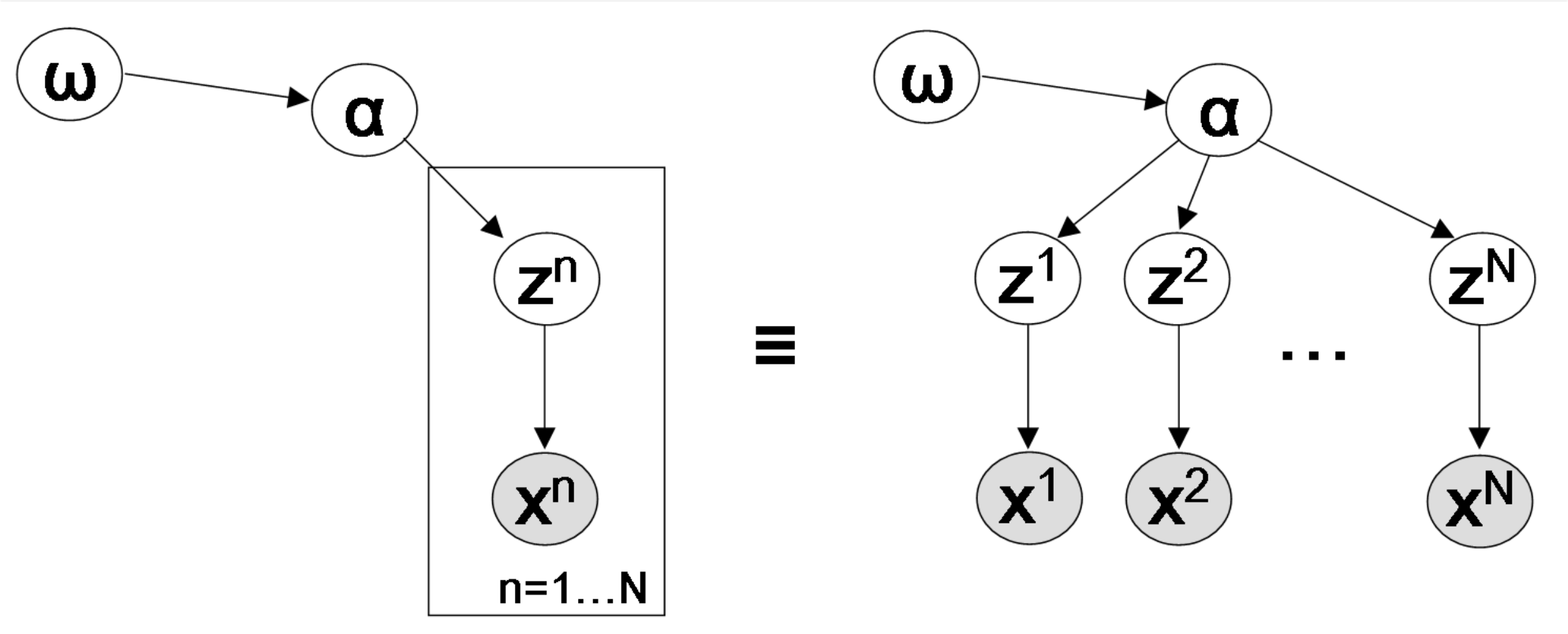}
\end{center}
\vspace{-1.2em}
\caption{Graphical model representation for BF-VAE-1 and BF-VAE-2: (Left) plate, (Right) unrolled version. The hyperparameter ${\boldsymbol\omega}$ is either $\{a_j\}$ (BF-VAE-1) or $\{r_j\}$ (BF-VAE-2).
}
\label{fig:gm_bfvae1}
\vspace{-0.5em}
\end{figure}

\section{Bayes-Factor-VAE (BF-VAE)}\label{sec:bfvae}

The key insight from \autoref{sec:analysis} is that, for relevant factors, it is necessary to have $p(z_j)$ different from  $\mathcal{N}(0,1)$. In this section we propose three different prior models to accomplish this goal in a principled Bayesian manners.

\subsection{Adjustable Gaussian Prior (BF-VAE-0)}\label{sec:bfvae0}

We first define a base model, also needed for subsequent more complex variations, which relaxes the fixed, identical variance assumption for priors $p(z_j)$:
\begin{equation}
p({\bf z}|{\boldsymbol\alpha}) = \prod_{j=1}^d p(z_j|\alpha_j)
  \ = \ \prod_{j=1}^d \mathcal{N}(z_j; 0, \alpha_j^{-1}),
\label{eq:bfvae0_prior}
\end{equation}
where ${\boldsymbol\alpha}>0$ are the precision parameters to be learned from data\footnote{Note that we fix the mean as $0$, and only learn the (inverse) variances $\alpha_j$. Although we can easily parametrize the mean as well, the form of \eqref{eq:bfvae0_prior} is equally flexible in terms of minimizing the KL, as shown in Supplement.}. 

We expect the learned $\alpha_j$ to be close to (apart from) $1$ for nuisance (relevant, resp.) $j$.  To explicitly express our preference of encouraging many dims $j$ to be nuisance, and avoid redundancy in the learned relevant variables, we add the regularizer, $(\alpha_j^{-1}-1)^2$, which leads to:
\begin{multline}
\min_{\theta,\nu,{\boldsymbol\alpha}} \ \ 
  \sum_{j=1}^d \mathbb{E}_{p_d({\bf x})} \Big[ \textrm{KL}( q(z_j|{\bf x}) || \mathcal{N}(z_j; 0, \alpha_j^{-1}) ) \Big]  \\
 + \textrm{Rec}(\theta,\nu) + \ \gamma \textrm{TC} 
\ + \ \eta \sum_{j=1}^d (\alpha_j^{-1}-1)^2.
\label{eq:obj_bfvae0}
\end{multline}

We denote this model by \textbf{BF-VAE-0}. The expected KL in \eqref{eq:obj_bfvae0} admits a closed form, resulting in added flexibility without extra computation, compared to e.g.,~\cite{factor_vae18}. Another benefit is the trade-off parameter $\eta$ acts as a proxy to control the cardinality of relevant factors; 
small $\eta$ encourages more relevant factors than large $\eta$. 


\subsection{Hierarchical Bayesian Prior  (BF-VAE-1)}\label{sec:bfvae1}

To extend BF-VAE-0 to a 
Bayesian hierarchical setting, in conjunction with \eqref{eq:bfvae0_prior}, we adopt a conjugate prior on 
$\bm{\alpha}$,
\begin{equation}
p({\boldsymbol\alpha}) = \prod_{j=1}^d p(\alpha_j)
  \ = \ \prod_{j=1}^d  \mathcal{G}(\alpha_j; a_j, b_j), \label{eq:bfvae1_alpha}
\end{equation}
where $\mathcal{G}(y; a,b) \propto y^{a-1} e^{-b y}$ is the Gamma distribution with parameters $a$ (shape) and $b$ (inverse scale) with $a,b>0$. 
We further set $b_j=a_j-1$, $a_j>1$, to express our preference for $\mathrm{Mode}[p(\alpha_j)] = 1$\footnote{
This preference also improved empirical performance. 
}. We let $\{a_j\}_{j=1}^d$ 
be the model parameters that can be learned from data. 
This model, named \textbf{BF-VAE-1}, has a graphical model representation shown in Fig.~\ref{fig:gm_bfvae1}.

A key aspect of this model is that by marginalizing out ${\boldsymbol\alpha}$, the prior $p({\bf z})$ becomes an infinite Gaussian mixture,  $p({\bf z}) = \int p({\boldsymbol\alpha}) \mathcal{N}({\bf z}; {\bf 0}, {\boldsymbol\alpha}^{-1}) d{\boldsymbol\alpha}$, a desideratum for relevant factors.   Because $\mathrm{Var}[p(\alpha_j)] \approx (a_j-1)^{-1}$, large $a_j$ will lead to $\lim_{a_j\rightarrow\infty} p(z_j|a_j) = \mathcal{N}({\bf z}; {\bf 0},{\bf 1})$, a nuisance factor.


We describe the variational inference for the model  where we introduce variational densities $q({\boldsymbol\alpha})$ and $q({\bf z}|{\bf x})$ to approximate the true posteriors as follows:
\begin{equation}
p({\boldsymbol\alpha},\{{\bf z}^n\}_{n=1}^N | \{{\bf x}^n\}_{n=1}^N ) \approx
    \overbrace{\prod_{j=1}^d \mathcal{G}(\alpha_j; \hat{a}_j, \hat{b}_j)}^{q({\boldsymbol\alpha})}
    \prod_{n=1}^N q({\bf z}^n | {\bf x}^n).
\label{eq:var_approx_bfvae1}
\end{equation}
This allows the average negative marginal data log-likelihood, $-\frac{1}{N}\log p(\{{\bf x}^n\})$, to be upper-bounded by\footnote{See Supplement for the derivations.}: 
\begin{multline}
\mathcal{U}_1 := 
\textrm{Rec}(\theta,\nu)
\ + \ \frac{1}{N} \textrm{KL}( q({\boldsymbol\alpha}) || p({\boldsymbol\alpha}) )\\
+ \mathbb{E}_{q({\boldsymbol\alpha})} \mathbb{E}_{p_d({\bf x})}\big[
      \textrm{KL}( q({\bf z}|{\bf x}) || p({\bf z}|{\boldsymbol\alpha}) ) \big].
\label{eq:elbo_bfvae1}
\end{multline}

$\textrm{Rec}(\theta,\nu)$ in \eqref{eq:elbo_bfvae1} is identical to that of VAE, 
while the other two admit closed forms; 
see Supplement for the details. 
The $\textrm{TC}$ term becomes an average over $q({\boldsymbol\alpha})$:
\begin{equation}
\textrm{TC}_1 := \mathbb{E}_{q({\boldsymbol\alpha})} \Bigg[
    \textrm{KL}(q({\bf z}|{\boldsymbol\alpha}) || \prod_{j=1}^d q(z_j|{\boldsymbol\alpha}) )
\Bigg],
\label{eq:tc_bfvae1}
\end{equation}
which turns out to be equal to $\textrm{TC}$ in \eqref{eq:tc}, 
since $q({\bf z}|{\boldsymbol\alpha}) := \int q({\bf z}|{\boldsymbol\alpha},{\bf x}) p_d({\bf x}) d{\bf x} = \int q({\bf z}|{\bf x}) p_d({\bf x}) d{\bf x} = q({\bf z})$.
The final optimization is then minimizing $(\mathcal{U}_1 + \gamma \textrm{TC}_1)$ wrt $(\theta, \nu)$ and $\{a_j, b_j, \hat{a}_j, \hat{b}_j\}_{j=1}^d$ with the constraint $b_j = a_j-1$.

BF-VAE-1 can capture the uncertainty in the precision parameters ${\boldsymbol\alpha}$ with no computational overhead  as all of the objective terms admit closed forms. Having learned the model from data $\mathcal{D}=\{{\bf x}^n\}_{n=1}^N$, the {\em data corrected prior},  $\overline{p}(z_j):=\int p(z_j|\alpha_j) p(\alpha_j|\mathcal{D}) d\alpha_j$, is approximated as:
\begin{equation}
\overline{p}(z_j) 
\approx \int p(z_j|\alpha_j) q(\alpha_j) d\alpha_j 
= t_{2\hat{a}_j}\bigg( z_j; 0, \frac{\hat{b}_j}{\hat{a}_j} \bigg),
\label{eq:corrected_prior}
\end{equation}
where $t_f(0,v)$ is the generalized Student's $t$ distribution with $\textrm{dof}$ $f$ and shape $v$. 
 $\overline{p}(z_j)$ informs us about the relevance of $z_j$: Large $\textrm{dof}$ implies nuisance (as the $t$ becomes close to Gaussian), while small suggests a relevant variable.
\subsection{Prior with Relevance Indicators   (BF-VAE-2)}\label{sec:bfvae2}

BF-VAE-1 allows only implicit control over the cardinality of relevant dims, assuming no explicit differentiation between relevant factors and nuisances. 
In this section we propose another model that can address these issues. 

The key idea\footnote{It is related to the well-known (Bayesian) variable selection problem~\cite{bayes_varsel09}, but clearly different in that the latter is typically framed within the standard regression setup where the variables (covariates) of interest are {\em observed} in the data. In our case, we aim to select the most relevant {\em latent} variables $z_j$'s that explain the major variation in the observed data.} 
is to introduce relevance indicator variables ${\bf r}\in[0,1]^d$ (high $r_j$ indicating relevance of $z_j$). We let ${\bf r}$ determine the shape of the hyper prior $p({\boldsymbol\alpha})$: If $r_j \approx 1$ (relevant), we make $p(\alpha_j)$ uninformative, thus $z_j$ far from $\mathcal{N}(0,1)$. In contrast, if $r_j \approx 0$ (nuisance), $p(\alpha_j)$ should strongly peak at $\alpha_j=1$, with $p(z_j)$ close to $\mathcal{N}(0,1)$. The following reparametrization of \eqref{eq:bfvae1_alpha} enables this control:
\begin{equation}
p({\boldsymbol\alpha}|{\bf r}) = 
\prod_{j=1}^d  \mathcal{G} \bigg(
    \alpha_j; \frac{1+2\epsilon}{r_j+\epsilon}, \frac{1+2\epsilon}{r_j+\epsilon}-1 \bigg),
\label{eq:bfvae2_alpha}
\end{equation}
where $\epsilon$ is a small positive number (e.g., $0.001$).

The indicator ${\bf r}$ naturally defines the relevant index set ${\bf R} = \{j: r_j \approx 1\}$), allowing us to decompose $q({\bf z})$ over ${\bf R}$ and ${\bf N}$ as $q({\bf z}_{\bf R}) \cdot \prod_{j \in {\bf N}} q(z_j)$\footnote{See Supplement for the derivations.}, 
making $\textrm{TC}$ into:
\begin{equation}
\textrm{KL}\Bigg( q({\bf z}_{\bf R}) || \prod_{j\in{\bf R}} q(z_j) \Bigg) 
\approx 
  \mathbb{E}_{q({\bf z}_{\bf R})} \bigg[ 
    \log \frac{D({\bf z}_{\bf R})}{1-D({\bf z}_{\bf R})} \bigg],
\label{eq:tc_r}
\end{equation}
focused only on relevant variables. Note that we suggest using the discriminator density ratio proxy, rhs of \eqref{eq:tc_r}, to evaluate TC, with $D(\cdot)$ optimized to discern samples from $q({\bf z}_{\bf R})$ from those of $\prod_{j\in{\bf R}} q(z_j)$. 

To turn \eqref{eq:tc_r} into a continuous space optimization problem, we rewrite $D({\bf z}_{\bf R})$ as $D({\bf r} \circ {\bf z})$, where $\circ$ is the element-wise (Hadamard) product, and introduce two additional regularizers to control the cardinality of ${\bf R}$ through $||{\bf r}||_1$ and the preference toward discrete values using the entropic prior $H({\bf r}) = -\sum_{j=1}^d \big( r_j \log r_j + (1-r_j) \log (1-r_j) \big)$.
This leads to the final objective:
\begin{equation}
\mathcal{U}_1 + 
\gamma \mathbb{E}_{q({\bf z})} \bigg[ 
\log \frac{D({\bf r} \circ {\bf z})}{1-D({\bf r} \circ {\bf z})} \bigg]  
 + \eta_S ||{\bf r}||_1 + \eta_H H({\bf r}),
\label{eq:obj_bfvae2}
\end{equation}
which is minimized over $(\theta,\nu)$, ${\bf r}$, and $\{\hat{a}_j,\hat{b}_j\}_{j=1}^d$, together with alternating gradient updates for $D(\cdot)$. 
%
In this model, named \textbf{BF-VAE-2}, the trade-off parameters $\eta_S$ and $\eta_H$ control the cardinality of relevant factors\footnote{We empirically demonstrate this in \autoref{sec:qualitative} and Supplement.
} large $\eta$ encourages few strong factors; for $\eta$ small, many weak factors could be learned. The learned relevance vector ${\bf r}$ can serve as an indicator discerning relevant factors from nuisances.

\section{Related Work
}\label{sec:related}

Most recent approaches to unsupervised disentanglement consider the learning objectives combining the VAE's loss in \eqref{eq:obj_vae} with regularization terms that encourage prior latent factor independence. 
%
In {\bf $\beta$-VAE}~\cite{beta_vae17}, the expected KL term of the VAE's objective is overemphasized, which can be seen as a proxy for the prior matching, i.e., minimizing $\textrm{KL}(q({\bf z})||p({\bf z}))$. In {\bf AAE}~\cite{aae16}, they aim to directly minimize the latter term via adversarial learning. As illustrated in our analysis in \autoref{sec:analysis}, the full independence of $q({\bf z})$ imposed in the TC, is important in the factor disentanglement, where the TC was estimated by the discriminator density ratio in {\bf Factor-VAE}~\cite{factor_vae18}, whereas {\bf TC-VAE}~\cite{tcvae} employed a weighted sampling strategy. Another alternative is the adversarial learning to minimize the Jensen-Shannon divergence in~\cite{nica17}, instead of KL in the TC. Quite closely related to the TC are:  {\bf DIP-VAE}~\cite{dip_vae18} that penalized the deviation of the variance 
of $q({\bf z})$ from identity, and  {\bf InfoGAN}~\cite{infogan16} that aimed to minimize the reconstruction error in the ${\bf z}$-space in addition to the reconstruction error in the ${\bf x}$-space. 

Recent deep representational learning attempts to extend the VAE by either adopting non-Gaussian prior models or partitioning latent variables into groups that are treated differently, both seemingly similar to our approach. In~\cite{hybrid_z}, a hybrid model that jointly represents discrete and continuous latents was introduced. In~\cite{lecun_disent}, under the partially labeled data setup, they separately treated the factors associated with the labels from those that are not, leading to a conditional factor model. The Gaussian prior assumption in VAE has been relaxed to allow more flexibility and/or better fit in specific scenarios. In {\bf VampPrior}~\cite{vampprior}, they came up with a reasonable encoder-based finite mixture model that approximates the infinite mixture model. In~\cite{vmf_vae} the von Mises-Fisher density was adopted to account for a hyper-spherical latent structure. The recent {\bf CHyVAE}~\cite{chyvae} employed the inverse-Wishart prior (generalization of Gamma), however, it mainly dealt with situations where latents can be correlated with one another apriori, via full prior covariance. The {\bf Hierarchical Factor VAE}~\cite{hfvae} instead focused on independence of groups of latent variables (group disentanglement). Although these recent works are closely related to ours, they either focused on different disentanglement goals, or extended the priors for inreased model capacity. 



\section{Evaluation}\label{sec:evaluation}

We evaluate our approaches\footnote{Our code is publicly available in https://seqam-lab.github.io/BFVAE/} on several benchmark datasets, where we assess the goodness of disentanglement both quantitatively and qualitatively. The former applies only to fully factor-labeled datasets, and we consider a comprehensive suite of disentanglement metrics in \autoref{sec:quantitative}. Qualitative assessment is accomplished through visualizations of data synthesis via latent space traversal. We also verify in \autoref{sec:qualitative} that the visually relevant/important aspects accurately correspond to those determined by the indicators we hypothesized in each of our three models. 



\textbf{1) Datasets.} We test all methods on the following datasets: \texttt{3D-Face}~\cite{3dfaces}, \texttt{Sprites}~\cite{dsprites} and its recent extension (\texttt{C-Spr})~\cite{color_dsprites} that fills the sprites with some random color (regarded as noise), \texttt{Teapots}~\cite{williams18}, and \texttt{Celeb-A}~\cite{celeba}. 
Also, we consider the subset of \texttt{Sprites} containing only the oval shape\footnote{Since the shape factor is in nature a discrete variable, the underlying models that assume continuous latent variables 
would be suboptimal. Instead of explicitly modeling a combination of discrete/continuous latent variables as in the recent hybrid model~\cite{hybrid_z}, we eliminate this discrete factor by considering only the oval-shape images only. 
}, denoted by \texttt{O-Spr}. 
The details of the datasets are described in the Supplement.
All datasets provide ground-truth factor labels except for \texttt{Celeb-A}. 
For all datasets, the image sizes are normalized to $64 \times 64$, and the pixel intensity/color values are scaled to $[0,1]$.  We use cross entropy as the reconstruction loss.

\textbf{2) Competing Approaches.}
We contrast our models with \textbf{VAE}~\cite{vae14}, \textbf{$\beta$-VAE}~\cite{beta_vae17}, and \textbf{F-VAE} (Factor-VAE)~\cite{factor_vae18}. 
We also compare our BF-VAE models with the recent \textbf{RF-VAE}~\cite{rfvae} that also considers differential treatment of relevant and nuisance latents. 

\textbf{3) Model Architectures and Optimization.} We adopt the model architectures and optimization parameters similar to those in~\cite{factor_vae18}. 
See Supplement for the details.


\subsection{Quantitative Results}\label{sec:quantitative}


We consider three disentanglement metrics\footnote{More details can be found in the Supplement.}: i) \textbf{Metric I}~\cite{factor_vae18} collects data samples with one ground-truth factor fixed with the rest randomly varied, encodes them as $\bm{z}$, finds the index of the latent with the smallest variance, and measures the accuracy of classification from that index to the ID of the fixed factor (the higher the better), ii) \textbf{Metric II}~\cite{rfvae} modifies Metric I by collecting samples of one factor varied with others fixed, and seeks the index of the largest latent variance. iii) \textbf{Metric III}~\cite{williams18} is based on regression from the latent vector to individual ground-truth factors, measuring three scores of prediction quality: {\em Disentanglement} for degree of dedication to each target, {\em Completeness} for degree of exclusive contribution by each covariate, and {\em Informativeness} for prediction error. Hence, higher scores are better for D and C, lower for I.

\autoref{tab:quant_results_all} summarizes all results, datasets and metrics. For all models across all datasets we use the latent dimension $d=10$.
Our models clearly outperform competing methods across all metrics in most instances. They are followed by RF-VAE, which also employs a notion of relevance, but not explicit non-Gaussianity. 


\begin{table*}[t]
\centering
\caption{Disentanglement metrics for benchmark datasets.  For Metric III, the three figures in each cell indicate Disentanglement / Completeness / Informativeness (top row based on the LASSO regressor, the bottom on the Random Forest. Note that the higher the better for D and C, while the lower the better for I. The best scores for each metric (within the margin of significance) among the competing models are shown in red and second-best in blue. 
}
\label{tab:quant_results_all}
\footnotesize
\vspace{+0.3em}
\begin{tabular}{|c|c||c|c|c|c|c|c|c|}
\hline
\multicolumn{2}{|c||}{Datasets/Metrics} & VAE & $\beta$-VAE & F-VAE & RF-VAE & BF-VAE-0 & BF-VAE-1 & BF-VAE-2 \\
\hline\hline
\multirow{4}{*}{\texttt{3D-Face}} & I & $\color[rgb]{1,0,0}100.0 \pm 0.0$ & $\color[rgb]{1,0,0}100.0 \pm 0.0$ & $\color[rgb]{1,0,0}100.0 \pm 0.0$ & $\color[rgb]{1,0,0}99.9 \pm 0.1$ & $\color[rgb]{1,0,0}100.0 \pm 0.0$ & $\color[rgb]{1,0,0}100.0 \pm 0.0$ & $\color[rgb]{1,0,0}100.0 \pm 0.0$ \\
\cline{2-9}
 & II & $\color[rgb]{0,0,1}93.4 \pm 0.7$ & $\color[rgb]{1,0,0}95.5 \pm 0.6$ & $\color[rgb]{0,0,1}92.8 \pm 1.1$ & $\color[rgb]{1,0,0}95.2 \pm 0.5$ & $\color[rgb]{1,0,0}95.6 \pm 0.5$  
 & $\color[rgb]{1,0,0}97.2 \pm 0.5$ & $\color[rgb]{1,0,0}97.5 \pm 0.5$ \\
\cline{2-9}
 & \multirow{2}{*}{III} & {\color[rgb]{0,0,1}.96} / .81 / {\color[rgb]{0,0,1}.37} & {\color[rgb]{0,0,1}.96} / .78 / .40 & {\color[rgb]{1,0,0}1.0} / {\color[rgb]{0,0,1}.82} / {\color[rgb]{1,0,0}.36} & {\color[rgb]{1,0,0}1.0} / {\color[rgb]{1,0,0}1.0} / .48 & {\color[rgb]{1,0,0}1.0} / {\color[rgb]{1,0,0}1.0} / .45  & {\color[rgb]{1,0,0}1.0} / {\color[rgb]{1,0,0}1.0} / .45 & {\color[rgb]{1,0,0}1.0} / {\color[rgb]{1,0,0}1.0} / .44 \\
 & & {\color[rgb]{0,0,1}.99} / .84 / {\color[rgb]{0,0,1}.26} & .98 / .86 / .31 & .96 / .83 / {\color[rgb]{1,0,0}.25} & {\color[rgb]{1,0,0}1.0} / {\color[rgb]{1,0,0}.93} / .37 & {\color[rgb]{1,0,0}1.0} / {\color[rgb]{0,0,1}.90} / .33 & {\color[rgb]{1,0,0}1.0} / {\color[rgb]{0,0,1}.90} / .34 & {\color[rgb]{1,0,0}1.0} / .88 / .41 \\
\hline\hline
\multirow{4}{*}{\texttt{Sprites}} & I & $80.2 \pm 0.3$ & $80.8 \pm 0.8$ & $81.9 \pm 1.0$ & $\color[rgb]{0,0,1}85.4 \pm 1.2$ & $\color[rgb]{0,0,1}87.9 \pm 0.9$ & $\color[rgb]{1,0,0}93.8 \pm 0.6$ & $\color[rgb]{0,0,1}85.5 \pm 0.8$ \\
\cline{2-9}
 & II & $58.2 \pm 1.4$ & $76.8 \pm 0.9$ & $77.6 \pm 1.4$ & $79.1 \pm 1.3$ & $\color[rgb]{0,0,1}82.7 \pm 1.1$ & $\color[rgb]{0,0,1}82.2 \pm 0.6$ & $\color[rgb]{1,0,0}85.9 \pm 1.2$ \\
\cline{2-9}
 & \multirow{2}{*}{III} & .59 / .68 / {\color[rgb]{1,0,0}.52} & .67 / .69 / {\color[rgb]{0,0,1}.53} & .84 / .84 / {\color[rgb]{0,0,1}.53} & .85 / .87 / {\color[rgb]{0,0,1}.53} & {\color[rgb]{0,0,1}.89} / {\color[rgb]{1,0,0}1.0} / .60 & {\color[rgb]{1,0,0}.92} /  {\color[rgb]{0,0,1}.90} / .54 & .88 / {\color[rgb]{1,0,0}1.0} / .58 \\
 & & .57 / .69 / .46 & .72 / {\color[rgb]{0,0,1}.84} / {\color[rgb]{0,0,1}.40} & {\color[rgb]{0,0,1}.73} / .82 / .41 & {\color[rgb]{0,0,1}.73} / .83 / .41 & {\color[rgb]{1,0,0}.75} / .83 / .44 & {\color[rgb]{1,0,0}.75} / .83 / {\color[rgb]{1,0,0}.34} & {\color[rgb]{1,0,0}.75} / {\color[rgb]{1,0,0}.86} / .48 \\
\hline\hline
\multirow{4}{*}{\texttt{C-Spr}} & I & $79.8 \pm 0.6$ & $81.2 \pm 0.4$ & $\color[rgb]{0,0,1}85.6 \pm 0.8$ & $80.7 \pm 0.9$ & $\color[rgb]{0,0,1}87.7 \pm 0.5 $ & $\color[rgb]{1,0,0}93.2 \pm 0.6$ & $\color[rgb]{1,0,0}94.7 \pm 0.8$ \\
\cline{2-9}
 & II & $61.2 \pm 1.5$ & $74.3 \pm 1.7$ & $76.2 \pm 0.8$ & $\color[rgb]{0,0,1}81.4 \pm 1.1$ & $\color[rgb]{1,0,0}83.0 \pm 1.4$ & $\color[rgb]{1,0,0}84.2 \pm 1.1$ & $\color[rgb]{1,0,0}83.5 \pm 0.7$ \\
\cline{2-9}
 & \multirow{2}{*}{III} & .52 / .55 / .54 & .77 / .82 / {\color[rgb]{0,0,1}.53} & .79 / .76 / {\color[rgb]{1,0,0}.52} & .87 / {\color[rgb]{0,0,1}.91} / .54 & {\color[rgb]{1,0,0}1.0} / {\color[rgb]{1,0,0}.95} / .56 & {\color[rgb]{0,0,1}.95} / {\color[rgb]{1,0,0}.95} / .58 & .86 / {\color[rgb]{0,0,1}.91}/ .56 \\
 & & .58 / .62 / .51 & .73 / .83 / {\color[rgb]{0,0,1}.39} & .75 / .83 / .42 & .64 /  .72 / {\color[rgb]{1,0,0}.30} & {\color[rgb]{1,0,0}.88} / .83 / .47 & .79 / {\color[rgb]{1,0,0}.88} / .42 & {\color[rgb]{0,0,1}.84} / {\color[rgb]{0,0,1}.85} / .45 \\
\hline\hline
\multirow{4}{*}{\texttt{O-Spr}} & I & $\color[rgb]{0,0,1}97.2 \pm 0.4$ & $75.3 \pm 0.6$ & $\color[rgb]{1,0,0}100.0 \pm 0.0$ & $\color[rgb]{1,0,0}100.0 \pm 0.0$ & $\color[rgb]{1,0,0}100.0 \pm 0.0$ & $\color[rgb]{1,0,0}100.0 \pm 0.0$ & $\color[rgb]{1,0,0}100.0 \pm 0.0$ \\
\cline{2-9}
 & II & $53.2 \pm 1.5$ & $70.2 \pm 1.2$ & $80.6 \pm 1.1$ & $95.4 \pm 0.5$ & $\color[rgb]{0,0,1}97.8 \pm 0.7$ & $\color[rgb]{1,0,0}99.8 \pm 0.2$ & $\color[rgb]{0,0,1}97.1 \pm 0.8$ \\
\cline{2-9}
 & \multirow{2}{*}{III} & .42 / .43 / .54 & {\color[rgb]{0,0,1}.58} / .49 / .49 & {\color[rgb]{1,0,0}1.0} / .88 / {\color[rgb]{1,0,0}.33} & {\color[rgb]{1,0,0}1.0} / {\color[rgb]{0,0,1}.99} / .49 & {\color[rgb]{1,0,0}1.0} / {\color[rgb]{1,0,0}1.0} / .42 & {\color[rgb]{1,0,0}1.0} / .97 / {\color[rgb]{0,0,1}.40} & {\color[rgb]{1,0,0}1.0}/ .93 / .42 \\
 & & .32 / .55 / .46 & .56 / .58 / .36 & .81 / .84 / .24 & .93 / .87 / {\color[rgb]{0,0,1}.22} & {\color[rgb]{1,0,0}.99} / {\color[rgb]{1,0,0}.93} / {\color[rgb]{0,0,1}.22} & {\color[rgb]{1,0,0}.99} / {\color[rgb]{0,0,1}.92} / {\color[rgb]{1,0,0}.21} & {\color[rgb]{0,0,1}.98} / .91 / .23 \\
\hline\hline
\multirow{4}{*}{\texttt{Teapots}} & I & $90.1 \pm 0.9$ & $56.9 \pm 1.1$ & $91.9 \pm 0.8$ & $\color[rgb]{1,0,0}98.7 \pm 0.4$ & $\color[rgb]{0,0,1}94.8 \pm 1.2$ & $\color[rgb]{1,0,0}97.6 \pm 0.3$ & $\color[rgb]{1,0,0}97.9 \pm 0.4$ \\
\cline{2-9}
 & II & $77.7 \pm 1.3$ & $47.3 \pm 0.9$ & $74.6 \pm 1.8$ & $\color[rgb]{0,0,1}83.1 \pm 1.2$ & $\color[rgb]{1,0,0}90.4 \pm 1.0$ & $\color[rgb]{0,0,1}82.7 \pm 1.3$ & $\color[rgb]{1,0,0}88.9 \pm 0.8$ \\
\cline{2-9}
 & \multirow{2}{*}{III} & .60 / .53 / .40 & .31 / .27 / .72 & .63 / .61 / .46 & .63 / .56 / {\color[rgb]{0,0,1}.37} & {\color[rgb]{1,0,0}.72} / .61 / {\color[rgb]{1,0,0}.34} & {\color[rgb]{0,0,1}.70} / {\color[rgb]{1,0,0}.65} / .48 & .67 / {\color[rgb]{0,0,1}.62} / .41 \\
 & & .81 / .72 / .31 & .45 / .61 / .52 & .75 / .78 / .29 & {\color[rgb]{1,0,0}.90} / {\color[rgb]{0,0,1}.79} / {\color[rgb]{0,0,1}.27} & {\color[rgb]{0,0,1}.89} / {\color[rgb]{1,0,0}.80} / {\color[rgb]{1,0,0}.25} & .78 / {\color[rgb]{1,0,0}.80} / .50 & .87 / {\color[rgb]{1,0,0}.80} / .32 \\
\hline
\end{tabular}
\end{table*}

\textbf{Comparison w/ High Capacity Priors.} 
Our analysis in \autoref{sec:analysis} states that a relevant dimension prior $p(z_j)$ needs to be non-Gaussian, flexible enough to match the aggregate posterior $q(z_j)$. Here, we consider an alternative prior with those properties.  Specifically, we use a F-VAE model with a Gaussian mixture prior $p({\bf z})= \sum_{k=1}^K \pi_k \mathcal{N}({\bf z}; {\boldsymbol\mu}_k; {\boldsymbol\Sigma}_k)$, with $\{(\pi_k,{\boldsymbol\mu}_k, {\boldsymbol\Sigma}_k)\}_k$ the model parameters to be optimized in conjunction with the F-VAE's parameters.  We contrast that model to our BF-VAE-2. %
\begin{figure}
\begin{center}
\includegraphics[trim = 3mm 2mm 3mm 2mm, clip, scale=0.235]{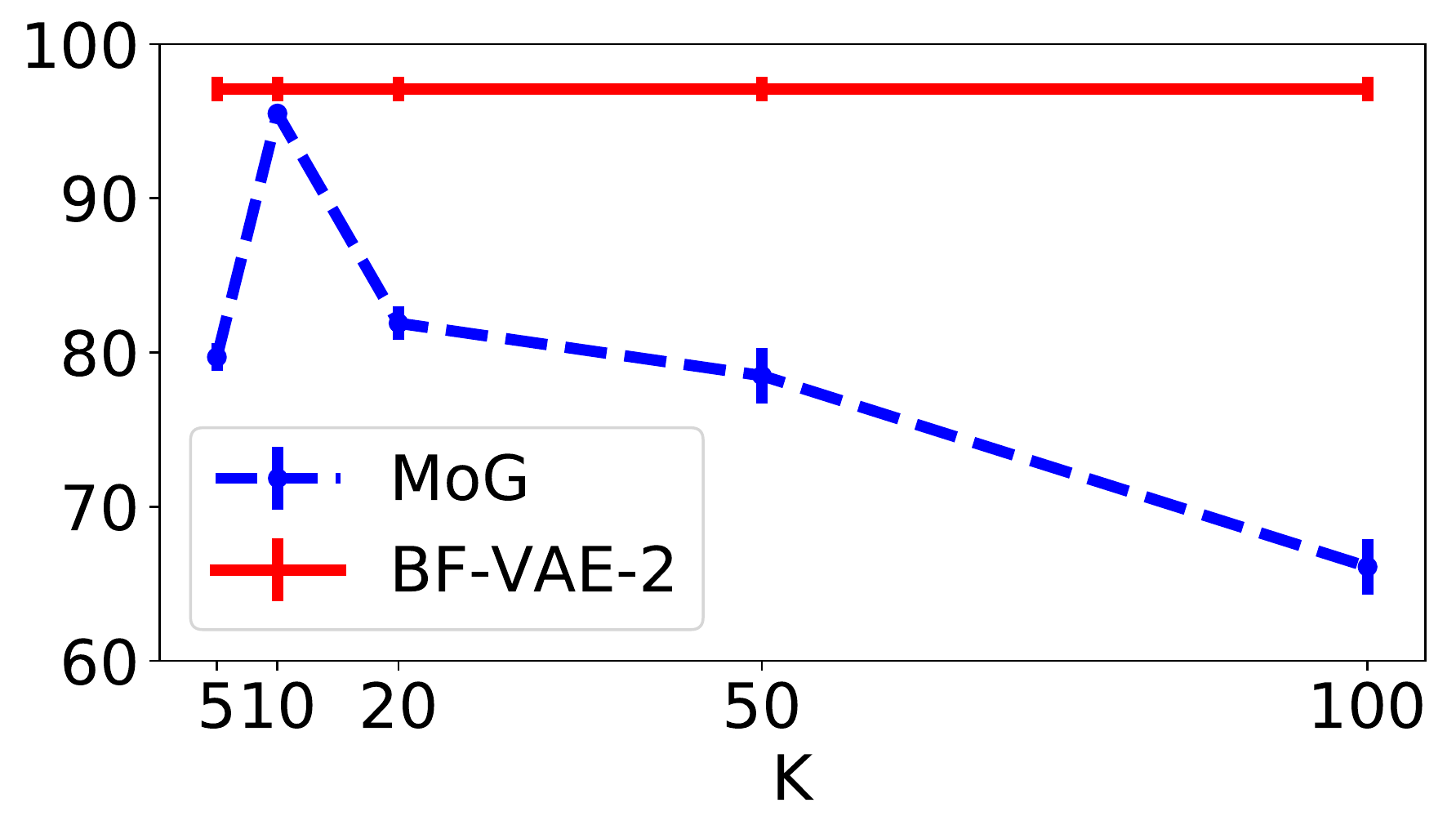}
\includegraphics[trim = 3mm 2mm 3mm 2mm, clip, scale=0.235]{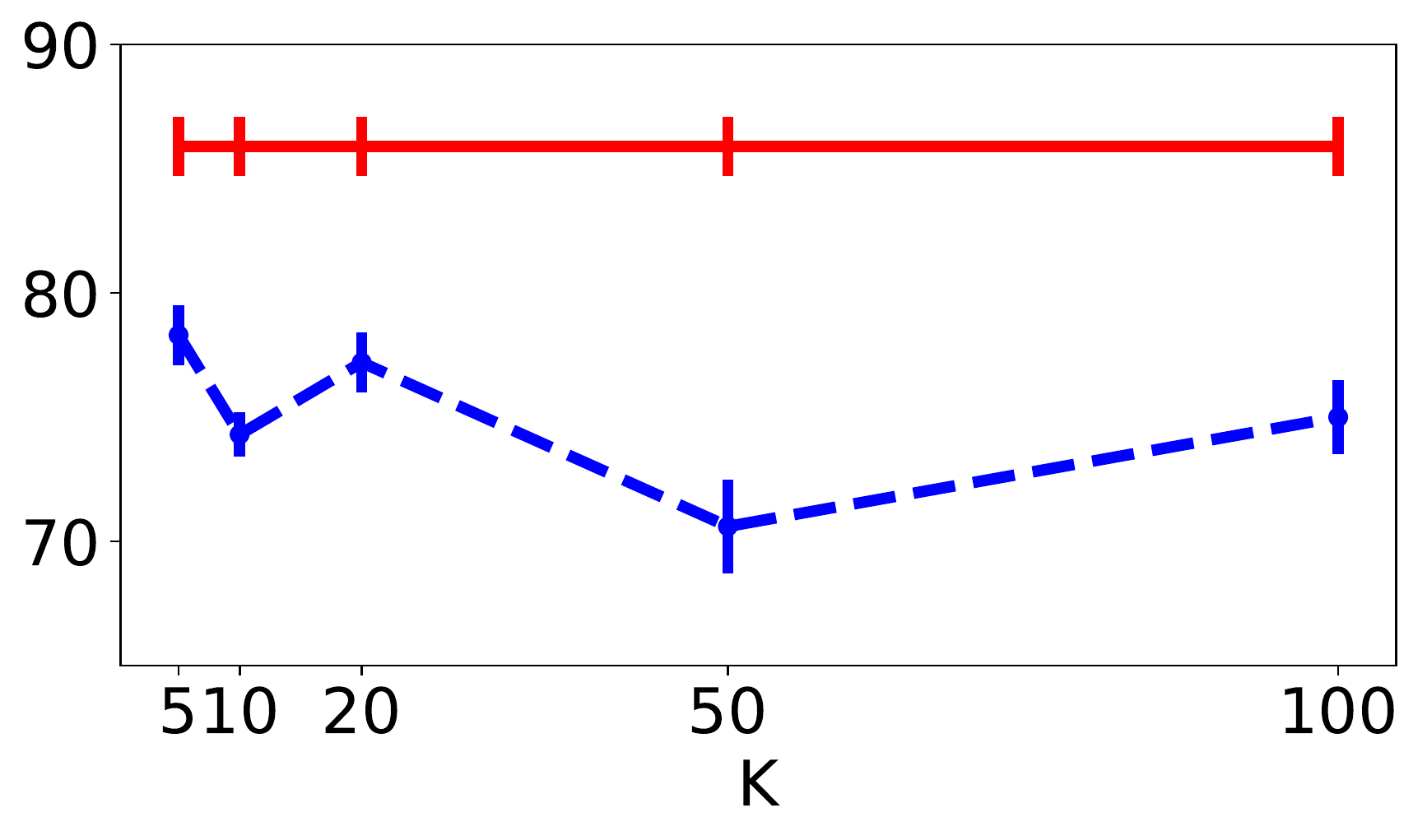}
\end{center}
\vspace{-1.2em}
\caption{Disentanglement performance (Metric II) of F-VAE with MoG prior (Blue/Dashed) with different mixture orders ($K$) vs.~BF-VAE-2 (Red/Solid) on \texttt{O-Spr} (Left) and \texttt{Sprites} (Right). 
}
\label{fig:mog_expmt}
\vspace{-0.5em}
\end{figure}
%
The disentanglement performances (Metric II scores) on \texttt{O-Spr} and \texttt{Sprites} are summarized in \autoref{fig:mog_expmt}, where we change the number of mixture components $K$ to control the degree of flexibility of the F-VAE mixture prior. Results show the high capacity mixture consistently underperforms our Bayesian model; as $K$ increases, it suffers from clear overfitting. This suggests the uncontrolled complex prior can be detrimental, in contrast to our controlled treatment of relevances.


\begin{figure}
\centering
%
\includegraphics[trim = 1mm 1mm 1mm 1mm, clip, scale=0.190]{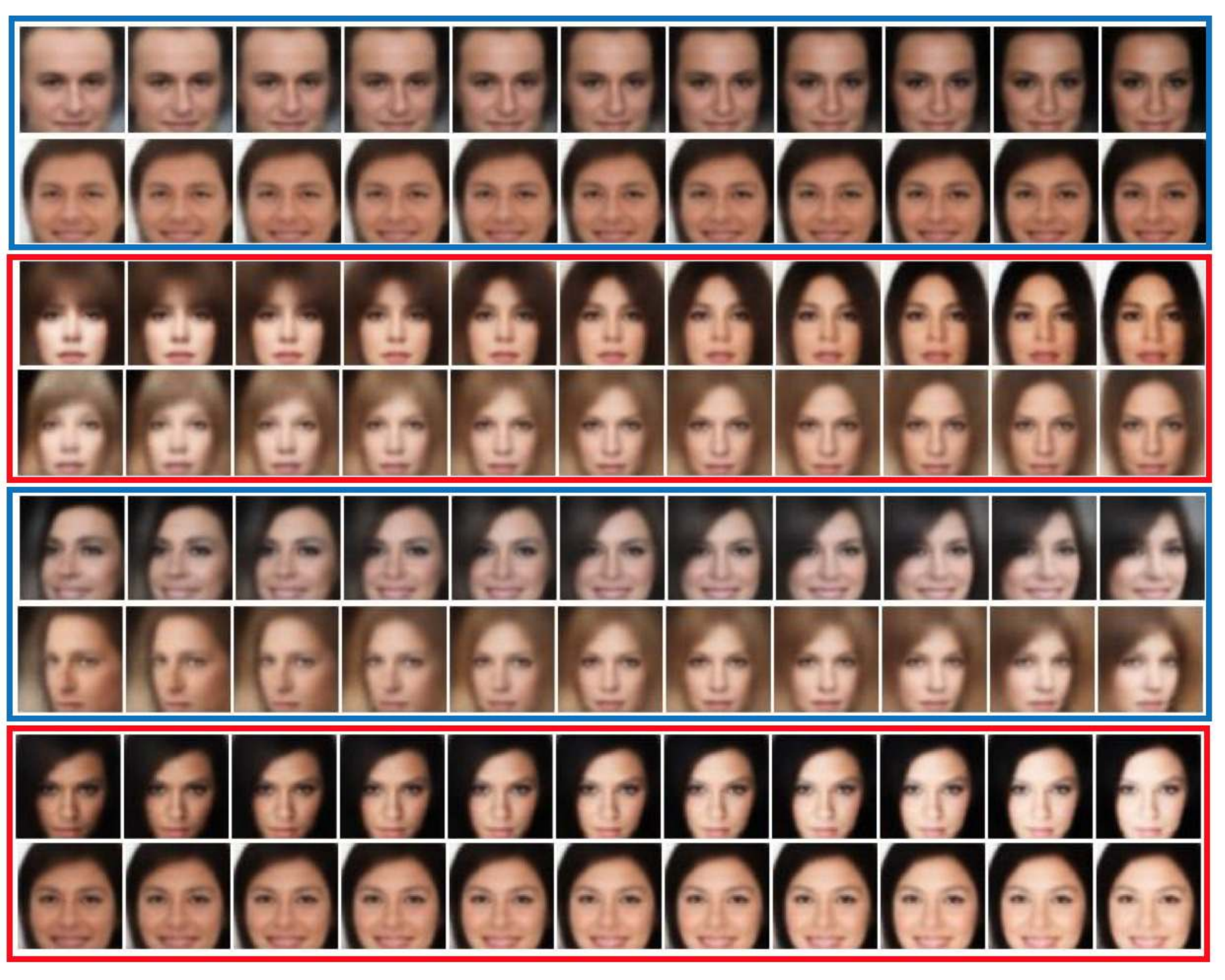} 
\includegraphics[trim = 1mm 0mm 1mm 1mm, clip, scale=0.190]{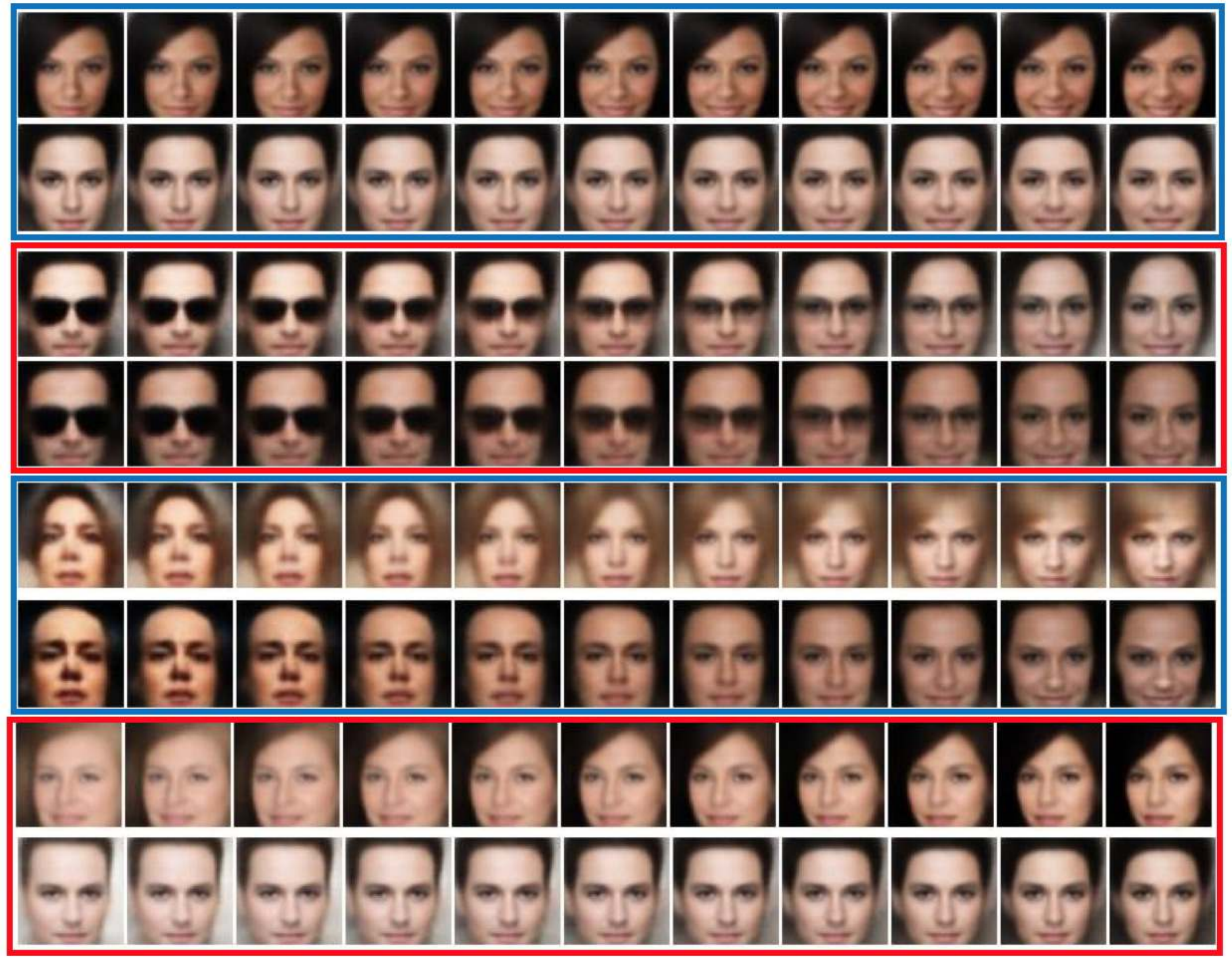}
%
\caption{Latent space traversal in BF-VAE-2 on \texttt{Celeb-A}. We train two BF-VAE-2 models with two different $\eta$ values ($\eta=\eta_S=\eta_H$ large and small). \textbf{(Left panel: strong factors)} contains latent traversal results with four latent variables (two subjects for each) that are detected (according to high $r_j$) by both $\eta$ small and large models. They correspond to (from top to bottom): gender, frontal hair, azimuth, and brightness, which are considered as strong/major factors. 
\textbf{(Right panel: weak factors)} shows traversal with four other latent variables that are detected (according to high $r_j$) only by the small $\eta$  model. They correspond to: smiling, sunglasses, elevation, and baldness, which are considered as weak/minor factors.
See Supplement for the enlarged images and further details.
}
\label{fig:bfvae2_trv_celeba}
\vspace{-0.5em}
\end{figure}

\begin{figure*}
\centering
\includegraphics[trim = 33mm 0mm 29mm 0mm, clip, scale=0.312]{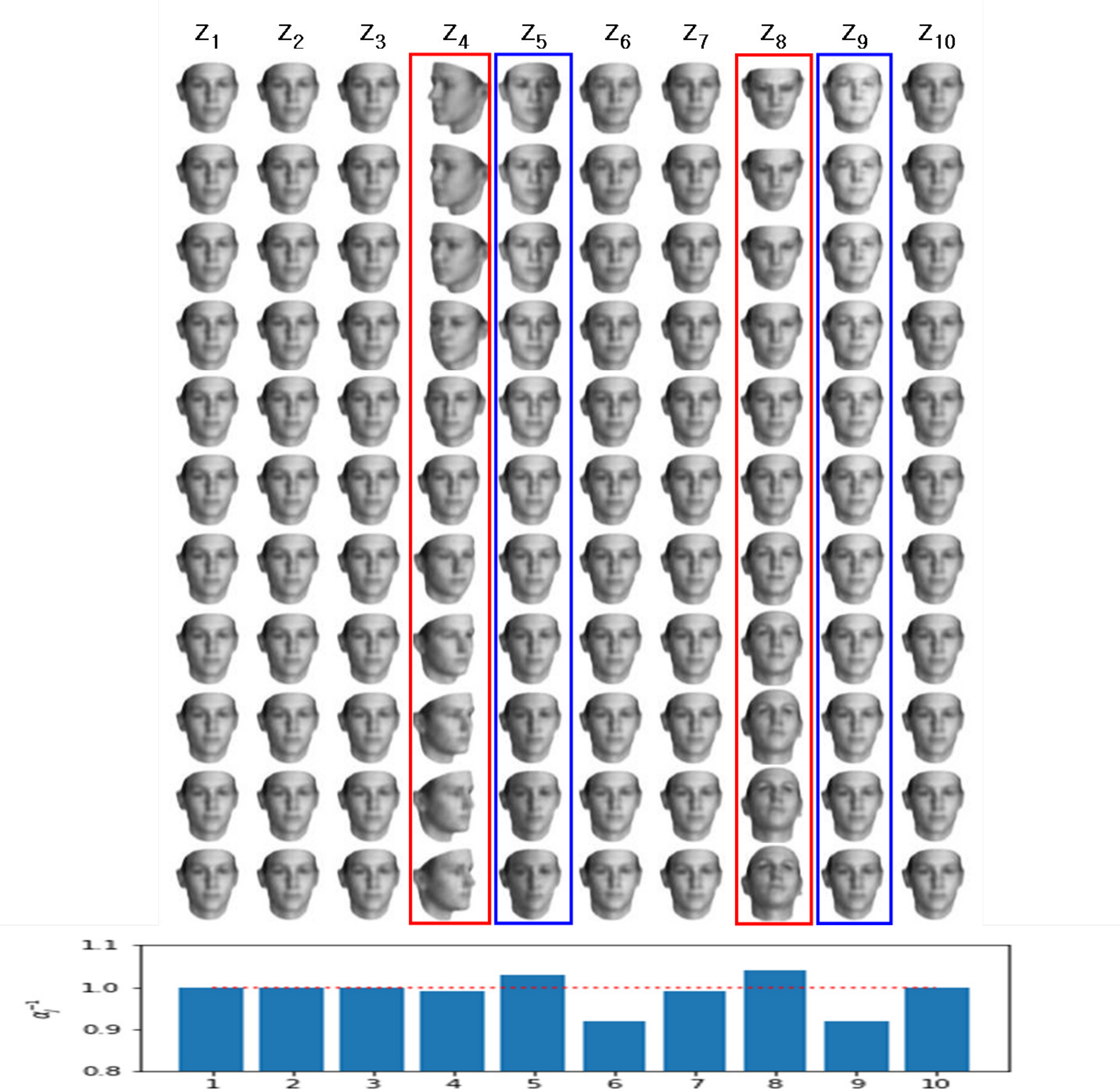} \ 
\includegraphics[trim = 33mm 0mm 29mm 0mm, clip, scale=0.312]{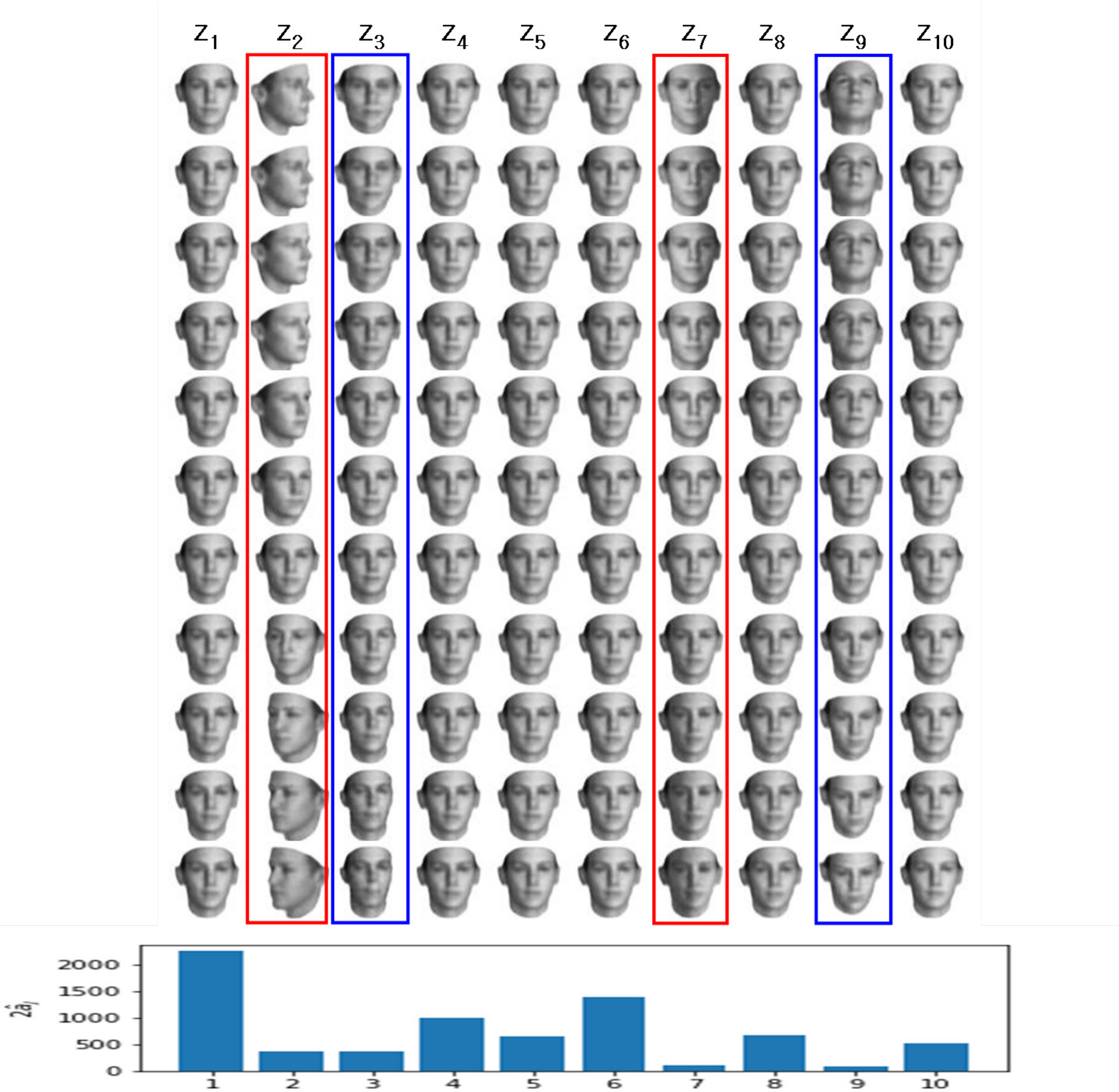} \ 
\includegraphics[trim = 33mm 0mm 29mm 0mm, clip, scale=0.312]{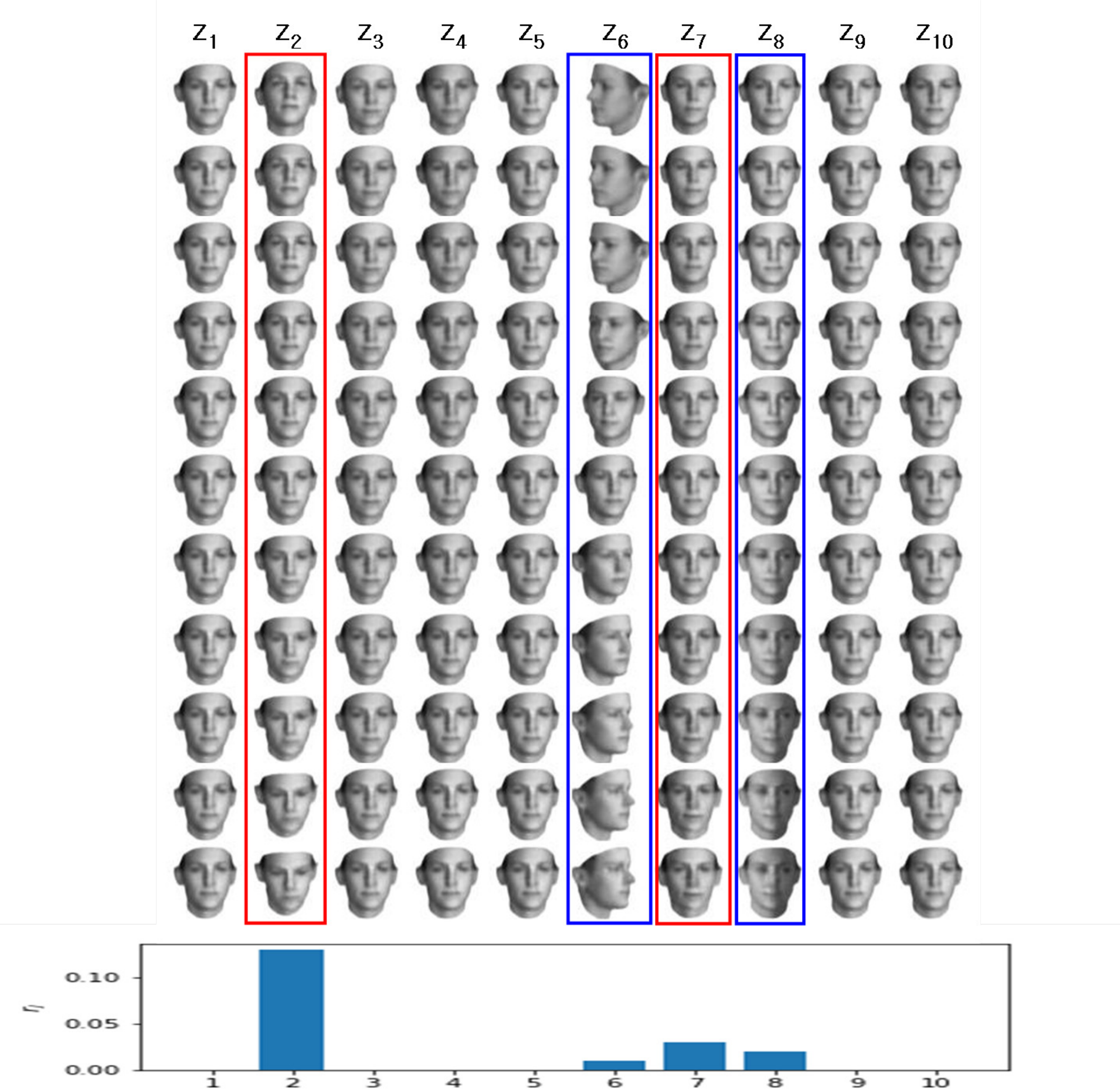} 
\caption{Latent space traversal in our three BF-VAE models on the \texttt{3D-Face} dataset. (Left) BF-VAE-0 with the learned prior variances ${\boldsymbol\alpha}^{-1}$ at the bottom (the value $1.0$ depicted as the red dotted line), (Middle) BF-VAE-1 with the DOF ($2\hat{a}_j$) of the corrected prior $\overline{p}(z_j)$ at the bottom, and (Right) BF-VAE-2 with the learned relevance vector ${\bf r}$ at the bottom. 
\textbf{(Left: BF-VAE-0)} The four  visually evident dimensions of variability ($z_4$, $z_5$, $z_8$, $z_9$) are highlighted within colored boxes, where each exactly matches one of the four ground-truth factors ($z_4=$ azimuth, $z_5=$ lighting, $z_8=$ elevation, and $z_9=$ subject ID). The learned $\alpha_j$ for all these four dims are away from $1$. 
\textbf{(Middle: BF-VAE-1)} The four  recovered, highlighted, dimensions match the ground-truth factors, and their $\overline{p}(z_j)$'s also have relatively small DOFs, as expected. 
\textbf{(Right: BF-VAE-2)} Again the four factors are nearly correctly identified, corresponding to the high values in the indicator variables $r_j$'s. 
}
\label{fig:trv_3dface}
\end{figure*}

\begin{figure*}
\centering
\includegraphics[trim = 33mm 0mm 29mm 0mm, clip, scale=0.312]{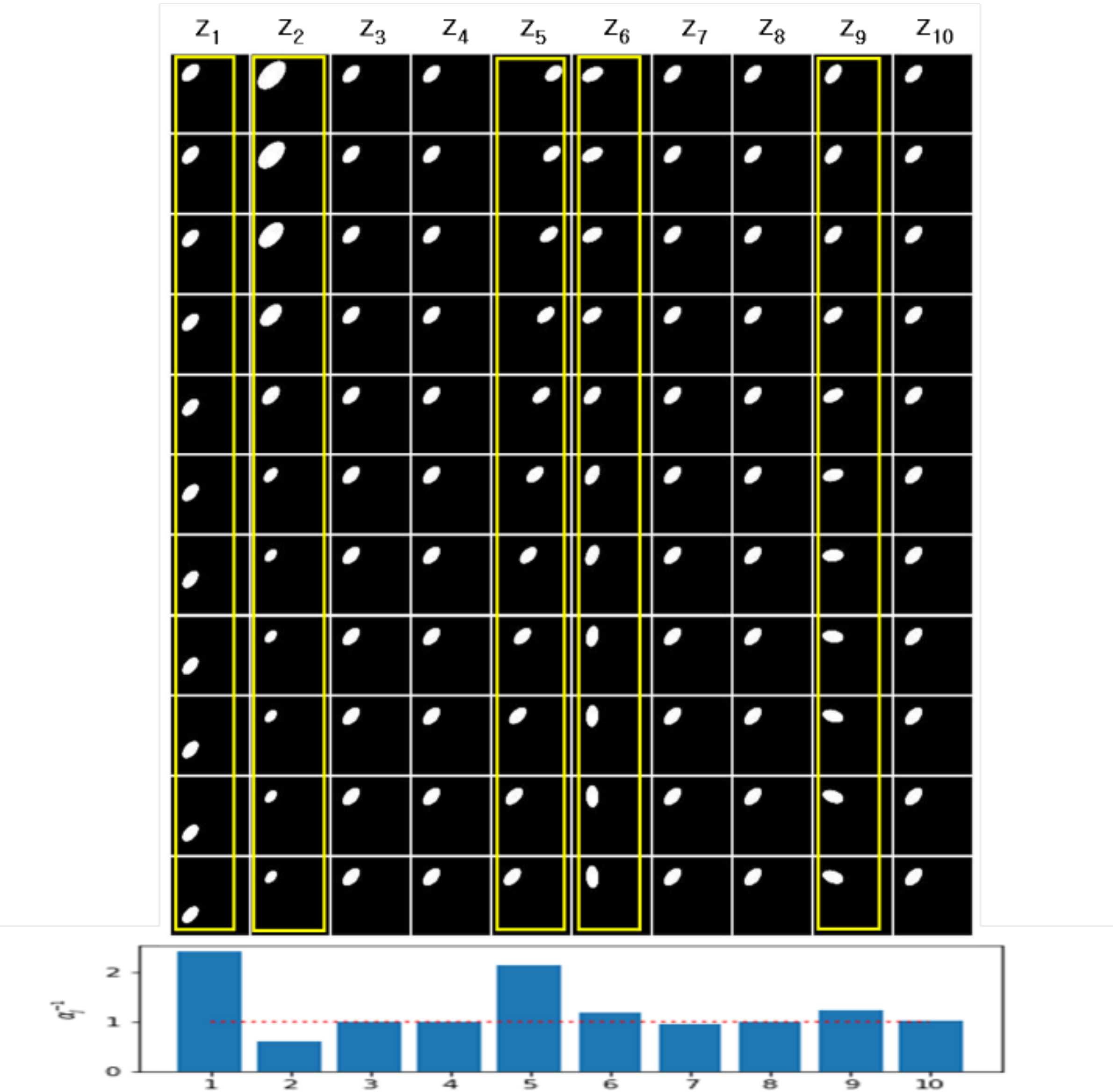} \ 
\includegraphics[trim = 33mm 0mm 29mm 0mm, clip, scale=0.312]{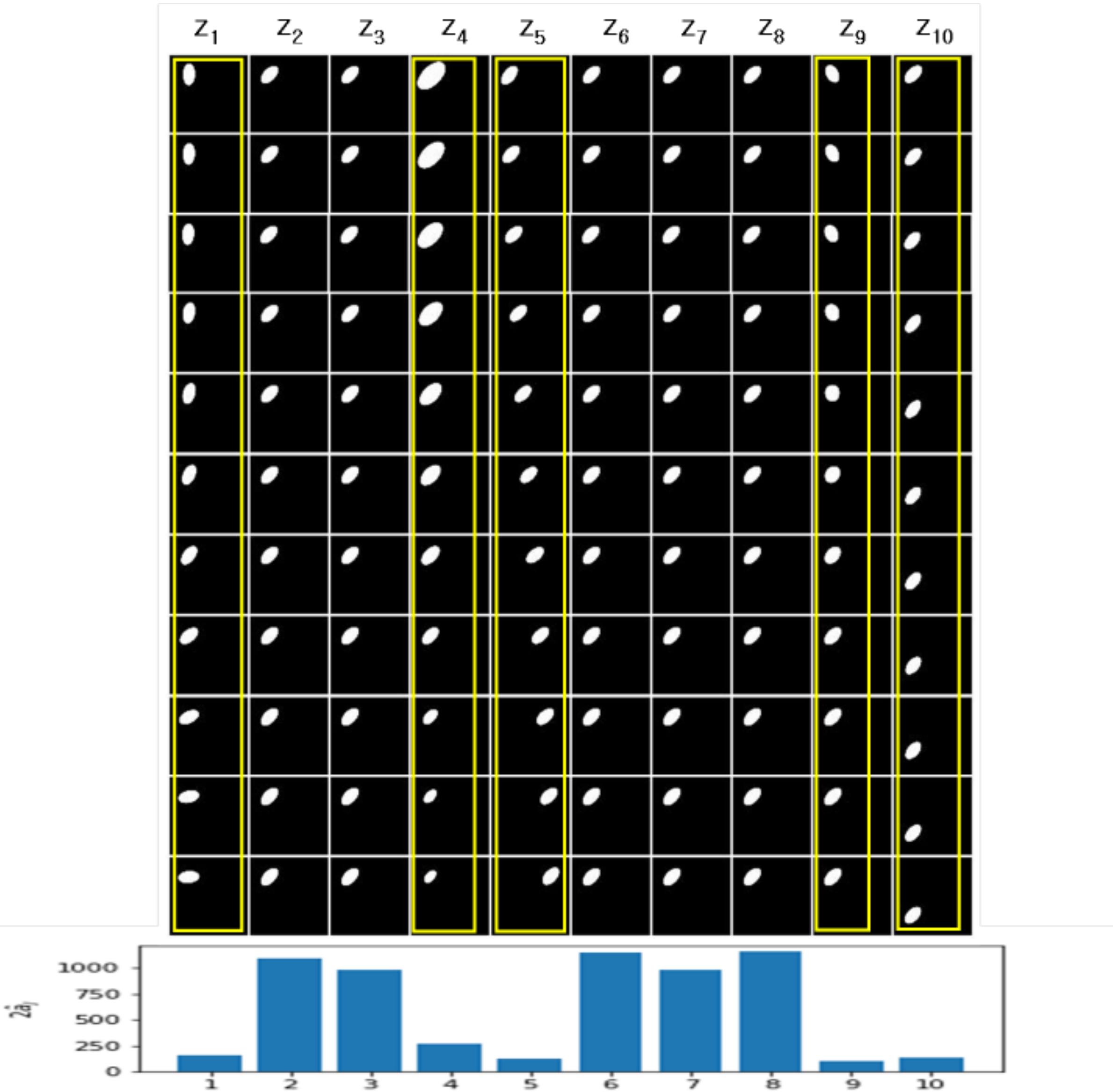} \ 
\includegraphics[trim = 33mm 0mm 29mm 0mm, clip, scale=0.312]{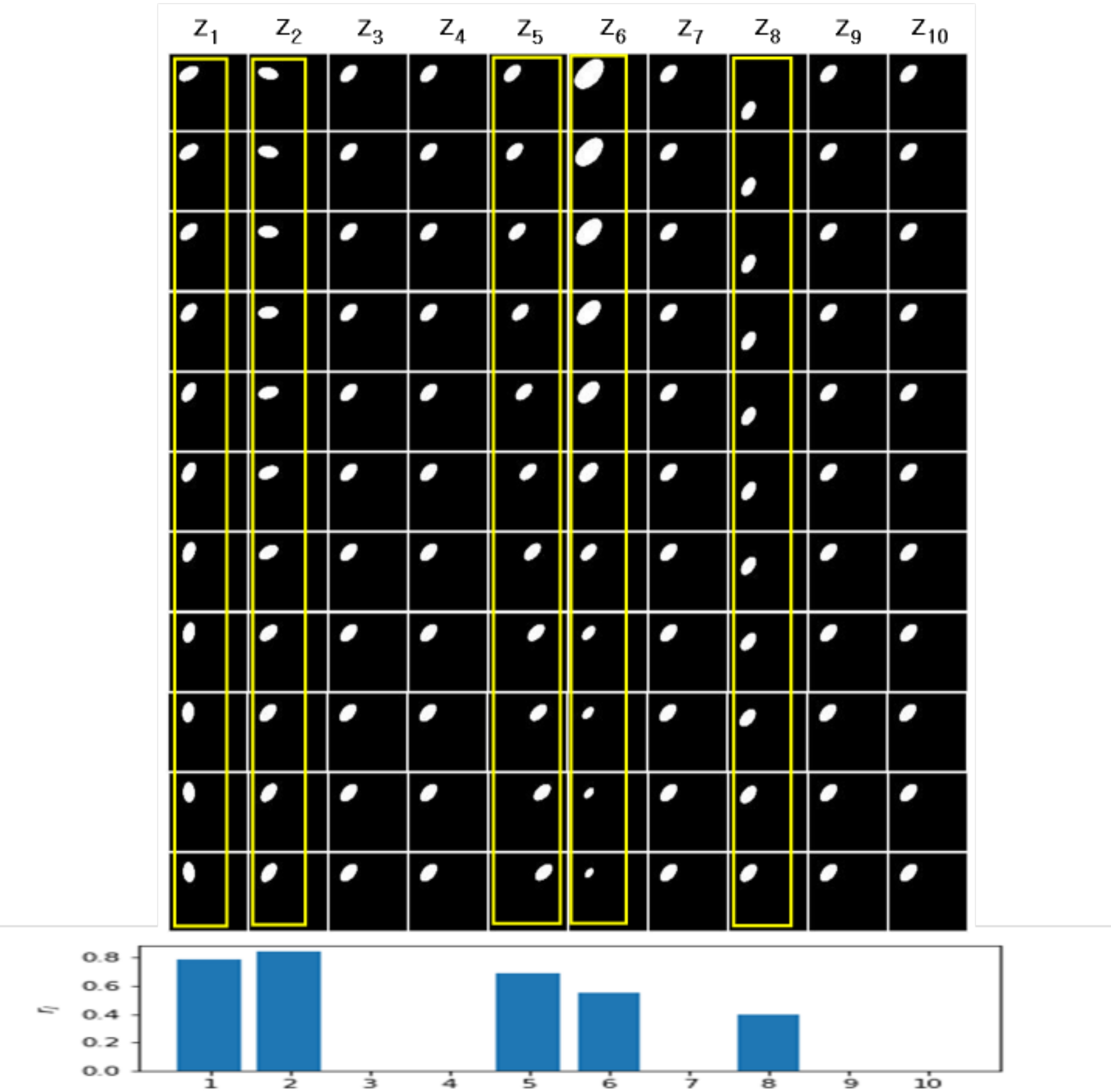} 
\caption{Latent 
traversal 
on \texttt{O-Spr}. 
The same interpretation as \autoref{fig:trv_3dface}. 
%
\textbf{(Left: BF-VAE-0)} Those five highlighted dimensions of major variability ($z_1$, $z_2$, $z_5$, $z_6$, $z_9$), match the four ground-truth factors (scale, X-, Y-pos, rotation), while the rotation is spread across $z_6$ and $z_9$. These factors also exactly correspond to the learned $\alpha_j$'s that are distant from $1$, as we anticipated. 
\textbf{(Middle: BF-VAE-1)} Similar to BF-VAE-0, it identifies five variables with the rotation spread across $z_1$ and $z_9$. These relevant variables, as expected, have small DOFs in $\overline{p}(z_j)$'s. 
\textbf{(Right: BF-VAE-2)} Again very similar to the previous two models. The learned ${\bf r}$ accurately indicates the relevant dimensions.
}
\label{fig:trv_oval}
\end{figure*}

\begin{figure*}
\centering
\includegraphics[trim = 33mm 0mm 29mm 0mm, clip, scale=0.312]{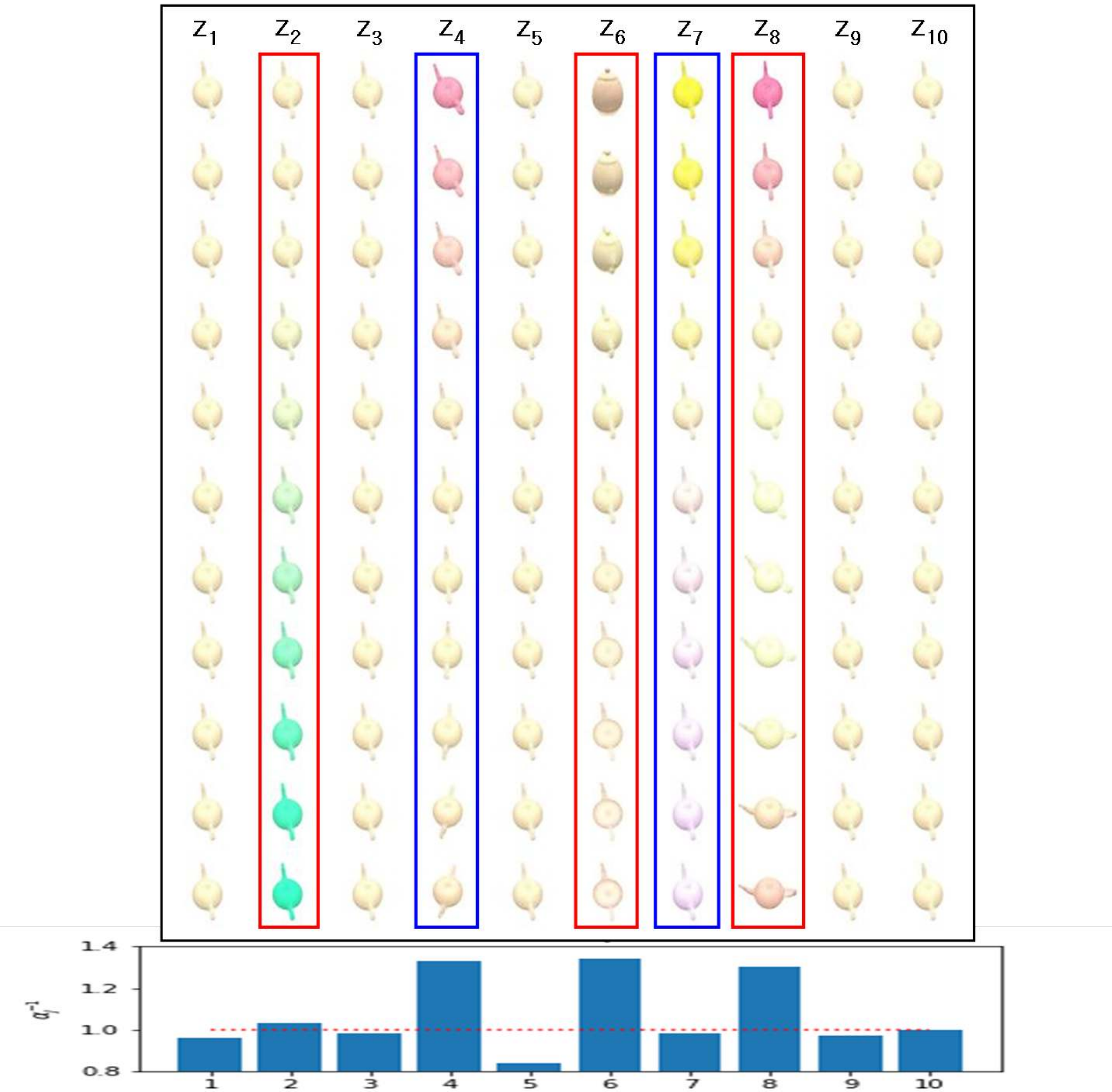} \
\includegraphics[trim = 33mm 0mm 29mm 0mm, clip, scale=0.312]{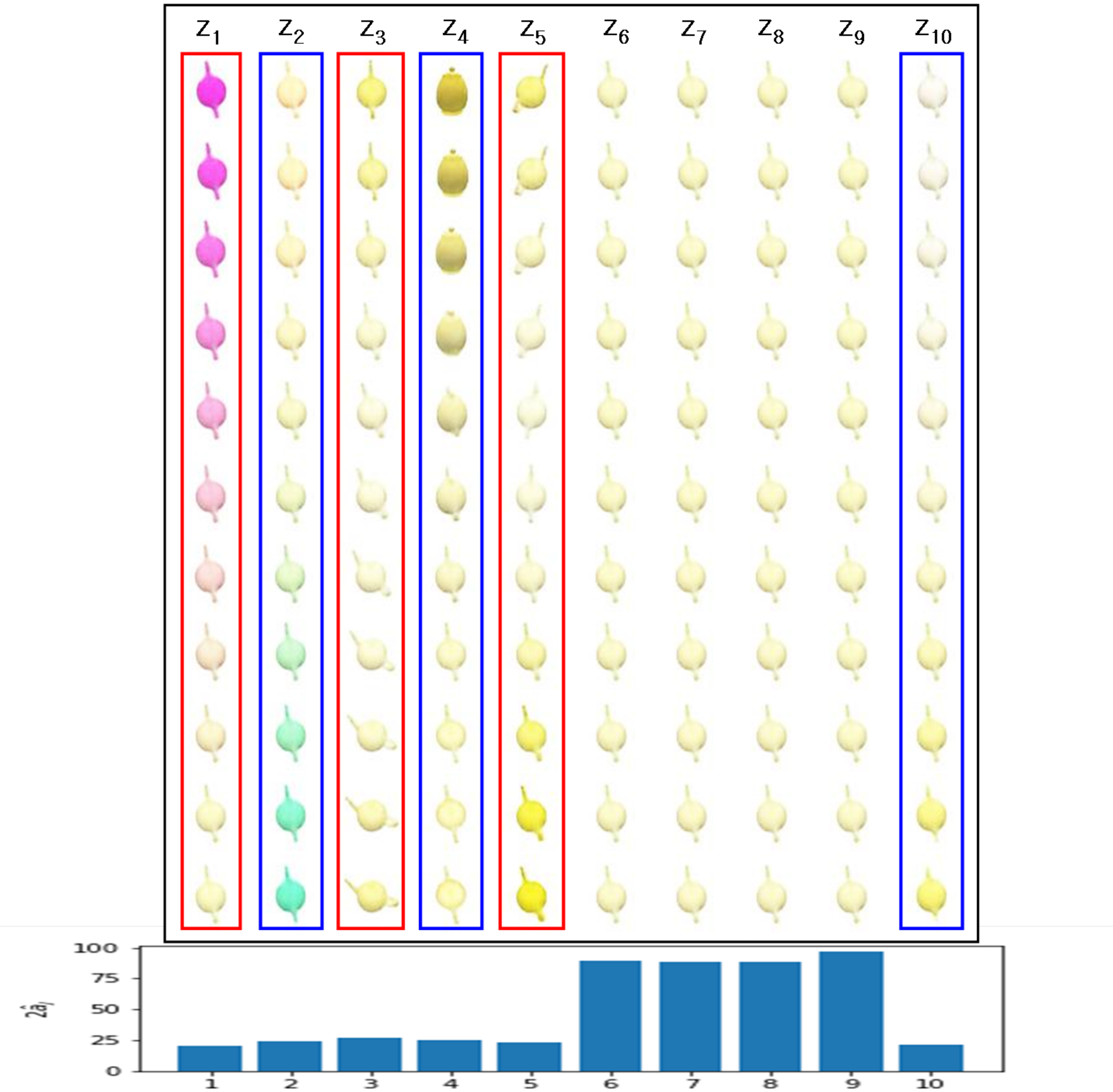} \ 
\includegraphics[trim = 33mm 0mm 29mm 0mm, clip, scale=0.312]{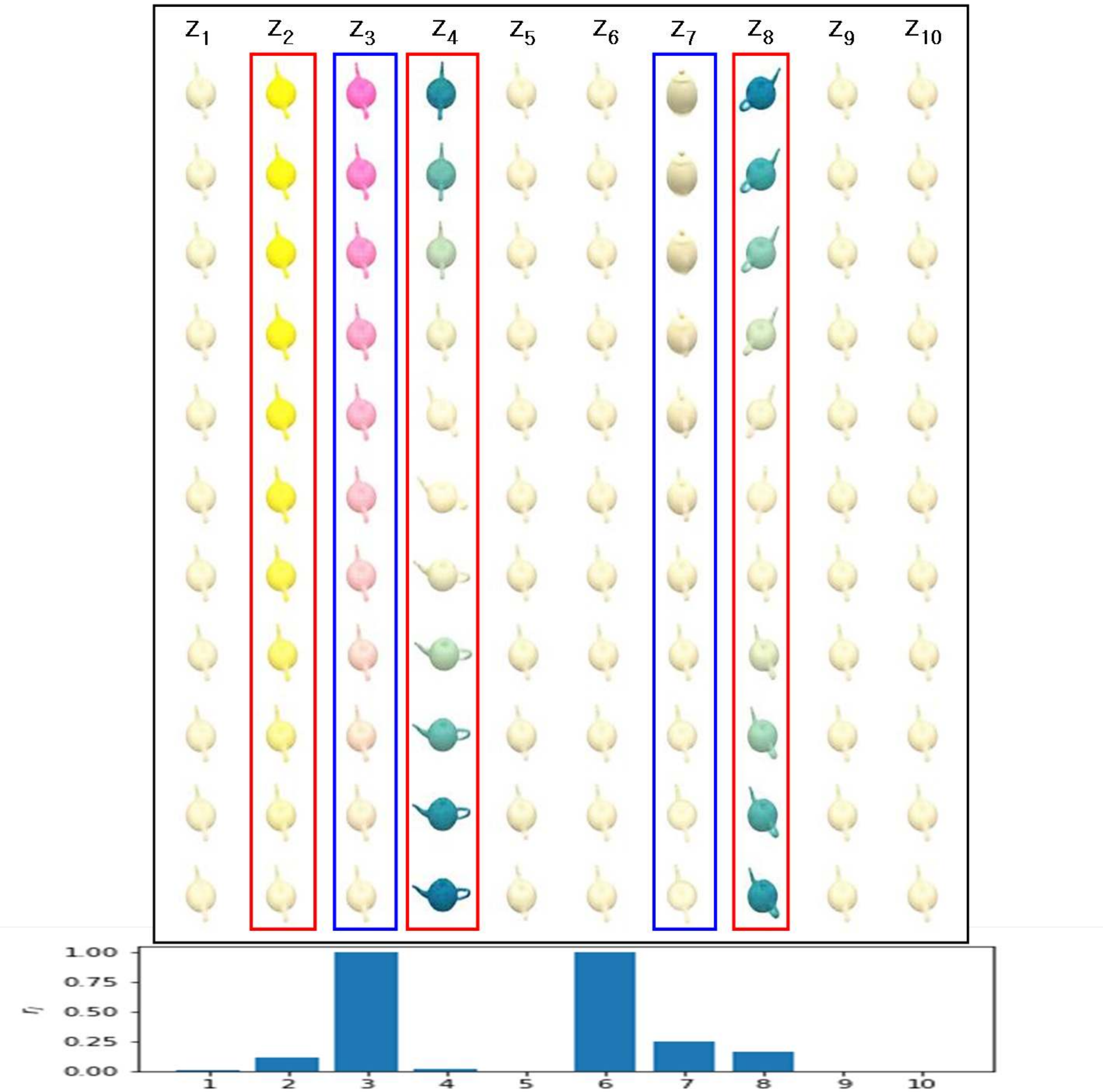} 
%
\caption{Latent traversal 
on \texttt{Teapots}.
The same interpretation as \autoref{fig:trv_3dface}. 
%
Note that the five ground-truth factors in this dataset are: two pose variations (azimuth and elevation) and three color changes (R,G,B).  
\textbf{(Left: BF-VAE-0)} The five variables that explain the major variability in images, ($z_2$, $z_4$, $z_6$, $z_7$, $z_8$), do not perfectly match the true factors one by one, and two or more factors are entangled in some variables (e.g., $z_8$ explains both color R and azimuth. Note that a similar failure was also observed in~\cite{williams18} with complex ResNet models.
\textbf{(Middle: BF-VAE-1)} and \textbf{(Right: BF-VAE-2)} Overall similar behaviors as BF-VAE-0, but the relevance indicators (implicit DOF in BF-VAE-1 and the explicit relevance vector ${\bf r}$ in BF-VAE-2) correctly identify the dimensions of major variability. 
}
\label{fig:trv_teapots}
\end{figure*}

\subsection{Qualitative Results}\label{sec:qualitative}

In this section we investigate qualitative performance of our BF-VAE approaches. We focus on: i) {\em Latent space traversal}: We depict images synthesized by traversing a single latent variable at a time while fixing the rest, and ii) {\em Accuracy of variable relevance indicator}: As discussed in \autoref{sec:bfvae}, our models have implicit/explicit indicators that point to relevant and nuisance variables. 
Specifically, i) BF-VAE-0 (learned $\alpha_j$): $j$ relevant if $\alpha_j$ is away from $1$, while $j$ is nuisance if $\alpha_j \approx 1$, ii) BF-VAE-1 (DOF of the corrected prior $\overline{p}(z_j)$, equal to $2\hat{a}_j$): 
$j$ is relevant if $\hat{a}_j$ is small (distant from Gaussian), and vice versa, 
iii) BF-VAE-2 (learned relevance indicator variable $r_j$): $j$ is relevant if $r_j$ is large, and vice versa.  

Due to the lack of space, we report selected results in this section, with more extensive results in Supplement. 
Results are shown for \texttt{3D-Face} (\autoref{fig:trv_3dface}), \texttt{O-Spr} (\autoref{fig:trv_oval}), and \texttt{Teapots} (\autoref{fig:trv_teapots}). The latent space traversal demonstrates that variation of each latent variable while the others are held fixed, visually leads to change in one of the ground-truth factors exclusively (except for the \texttt{Teapots}).  Also, these visually identified factors indeed correspond to those variables indicated as relevant by our models. See the figure captions for details.

\textbf{Control of Cardinality of Relevant Factors.} One of the distinguishing benefits of our BF-VAE-2 (and also BF-VAE-0) is that the trade-off parameter(s) $\eta$ can control the number of relevant factors to be detected by the model. We visually verify this on \texttt{Celeb-A} dataset. As shown in  \autoref{fig:bfvae2_trv_celeba} (detailed in the caption), adopting large $\eta$ leads only strong factors to be detected, while having small $\eta$ allows many weak factors identified.

\section{Conclusion}

This work demonstrates that, for recovery of disentangled factors of variation in data, it is essential to embrace and model the non-Gaussian nature of relevant factors while, at the same time, discerning them from Gaussian nuisances, in contrast to traditional prior assumptions used for VAEs.  We showed that a VAE endowed with a hierarchical Bayesian  prior, the BF-VAE, can effectively model both aspects of this task.  Empirical evaluation on benchmark datasets validates this ability of the BF-VAE family, showing consistently leading performance across three disentanglement metrics.  We also demonstrated the models' ability to recover strong indicators of data variability, with clear qualitative effects observed through traversals in the learned factor space and re-synthesis of data via the models' learned decoding stage.


{\small
\bibliographystyle{ieee_fullname}
\bibliography{bfvae}
}

\end{document}